\documentclass{elsarticle}
\usepackage[ruled,linesnumbered]{algorithm2e}
\usepackage{epstopdf}
\usepackage{amsmath}
\usepackage{amsfonts}
\usepackage{graphicx}
\usepackage{subfigure}
\usepackage{multirow}
\usepackage{booktabs}
\usepackage{tabularx}
\usepackage{float}
\usepackage[colorlinks,
            linkcolor=blue,
            anchorcolor=blue,
            citecolor=blue]{hyperref}
\usepackage{diagbox}

\journal{Swarm and Evolutionary Computation}

\UseRawInputEncoding
\begin{document}

\begin{frontmatter}

\title{A Distribution Evolutionary Algorithm for the Graph Coloring Problem}
\tnotetext[mytitlenote]{Y. Xu and H. Cheng contributed equally to this research.}


\author[WUT]{Yongjian Xu}
\ead{3414727121@qq.com}

\author[WSU]{Huabin Cheng}
\ead{2020111132@wsyu.edu.cn}

\author[WUT1]{Ning Xu}
\ead{xuning@whut.edu.cn}

\author[WUT]{Yu Chen\corref{correspondingauthor}}
\cortext[correspondingauthor]{Corresponding authors}
\ead{ychen@whut.edu.cn}

\author[SCNU]{Chengwang Xie\corref{correspondingauthor}}
\ead{chengwangxie@m.scnu.edu.cn}
\address[WUT]{School of Science, Wuhan University of Technology, Wuhan, 430070, China}
\address[WSU]{Department of Basic Science, Wuchang Shouyi University, Wuhan, 430064, China}
\address[WUT1]{School of Information Engineering, Wuhan University of Technology, Wuhan, 430070, China}
\address[SCNU]{School of Data Science and Engineering, South China Normal University, Guangdong, 516600, China}

\begin{abstract}
{\color{blue}Graph coloring is a challenging combinatorial optimization problem with a wide range of applications. In this paper, a distribution evolutionary algorithm based on a population of probability model (DEA-PPM) is developed to address it efficiently. Unlike existing estimation of distribution algorithms where a probability model is updated by generated solutions, DEA-PPM employs a distribution population based on a novel probability model, and an orthogonal exploration strategy is introduced to search the distribution space with the assistance of an refinement strategy. By sampling the distribution population, efficient search in the solution space is realized based on a tabu search process. Meanwhile, DEA-PPM introduces an iterative vertex removal strategy to improve the efficiency of $k$-coloring, and an inherited initialization strategy is implemented to address the chromatic problem well. The cooperative evolution of the distribution population and the solution population leads to a good balance between exploration and exploitation. Numerical results demonstrate that the DEA-PPM of small population size is competitive to the state-of-the-art metaheuristics.}
\end{abstract}


\end{frontmatter}


\section{Introduction}\label{in}
Given an undirected graph $G=(V,E)$ with a vertex set $V$ and a edge set $E$, the (vertex) graph coloring problem (GCP) assigns colors to vertexes such that no adjacent vertexes share the same color. If $G$ can be colored by $k$ different colors without color conflicts, it is \emph{$k$-colorable}. The smallest value of color number $k$ such that $G$ is $k$-colorable is its \emph{chromatic number}, denoted by $\chi (G)$. There are two instances of the GCP, the \emph{$k$-coloring problem} attempting to color a graph with $k$ colors and the \emph{chromatic number problem} trying to get the chromatic number of $G$, both of which are extensively applied in scientific and engineering fields. {\color{red}Due to the NP-completeness of GCPs, some relaxation methods were proposed to transform the combinatorial GCPs to continuous optimization problems~\cite{Greenwood2009UsingDE,artacho2020enhanced,Goudet2021PopulationbasedGD}. However, the transformation will lead to continuous problems with distinct landscapes, and global optimal solutions of the original GCPs could be quite different from those of the relaxed problems.

Accordingly, a variety of metaheuristics have been developed to address the original GCPs efficiently~\cite{MOSTAFAIE2020104850}. \emph{Individual-based metaheuristics} search the solution space by  single-point iteration schemes, contributing to their fast convergence and low complexity~\cite{galinier2006survey}. However, their performance relies heavily on the initial solution and the definition of neighborhood, which makes it challenging to balance the exploration and the exploitation.   \emph{Population-based metaheuristics} perform cooperative multi-point search in the solution space, but a comparatively large population is usually necessary for the efficient search in the solution space, which makes it inapplicable to large-scale GCPs~\cite{MOSTAFAIE2020104850}.

Recently, metaheuristics based on probability models have been widely employed to solve complicated optimization problems~\cite{Dorigo2006,Hauschild2011,Xiong2018}. As two popular instances,
the ant colony optimization (ACO)~\cite{Dorigo2006} and the estimation of distribution algorithm (EDA)~\cite{Hauschild2011} employ a single probability model that is gradually updated during the iteration process, which makes it difficult to balance the global exploration and the local exploitation in the distribution space. The quantum-inspired evolutionary algorithm (QEA) performs an active update of probability model by the Q-gate rotation, whereas it is a kind of local exploitation that cannot explore the distribution space efficiently~\cite{Xiong2018}. To remedy the aformentioned issues, we propose a distribution evolutionary algorithm based on a population of probability model (DEA-PPM), where a balance between the convergence performance and the computational complexity could be kept by evolution of small populations.} {\color{blue} Contributions of this work are as follows.
\begin{itemize}
  \item We propose a novel distribution model that incorporates the advantages of EDAs and QEAs.
  \item Based on the proposed distribution model, an orthogonal exploration strategy is introduced to search the probability space with the assistance of a tailored refinement strategy.
  \item For the chromatic problem, an inherited initialization is presented to accelerate the convergence process.
\end{itemize}
}

{\color{red} Rest of this paper is organized as follows. Section \ref{RW} presents a brief review on related works. The proposed distribution model is presented in Section \ref{DisMod}, and Section \ref{DEAPPM} elaborates details of DEA-PPM.  Section \ref{result} investigates the influence of parameter and the distribution evolution strategies, and the competitiveness of DEA-PPM is verified by numerical experiments. Finally, we summarize the work in Section \ref{con}.}

\section{Literature Review}\label{RW}
\subsection{Individual-based metaheuristics for GCPs}

Besides the simulated annealing~\cite{titiloye2011quantum,PAL2012321} and  the variable neighborhood search~\cite{avanthay2003variable}, the tabu search (TS) is one of the most popular individual-based metaheuristics applied to solve the GCPs~\cite{Hertz1987UsingTS}. Porumbel \emph{et al.}~\cite{porumbel2013informed} improved the performance of TS by evaluation functions that incorporates the structural or dynamic information in addition to the number of conflicting edges.
Bl\"{o}chliger and Zufferey~\cite{BLOCHLIGER2008960} proposed a TS-based constructive strategy, which constructs feasible but partial solutions and gradually increases its size to get the optimal color assignment of a GCP.
Hypothesizing that high quality solutions of GCPs could be grouped in clusters within spheres of a specific diameter, Porumbel \emph{et al.}~\cite{porumbel2010search} proposed two improved TS variants using a learning process and a tree-like structure of the connected spheres. Assuming that each vertex only interacts with a limited number of components, Gal\'{a}n~\cite{Galn2017SimpleDG} developed a decentralized coloring algorithm, where colors of vertexes are modified according to those of the adjacent vertexes to iteratively reduce the number of edge conflicts. Sun \emph{et al.}~\cite{Sun2021ASM} established a solution-driven multilevel optimization framework for GCP, where an innovative coarsening strategy that merges vertexes based on the solution provided by the TS, and the uncoarsening phase is performed on obtained coarsened results to get the coloring results of the original graph.
To color vertexes with a given color number $k$, Peng \emph{et al.}~\cite{peng2021vcolor} partitioned a graph  into a set of connected components and a vertex cut component, and combined the separately local colors by an optimized maximum matching based method.

Since a probability model can provide a bird's-eye view for the landscape of optimization problem, Zhou \emph{et al.}~\cite{Zhou2016ReinforcementLB,Zhou2018ImprovingPL} proposed to enhance the global exploration of individual-based metaheuristics by the introduction of probability models.
They deployed a probabilistic model for the colors of vertexes, which is updated with the assistance of a reinforcement learning technology based on discovered local optimal solutions~\cite{Zhou2016ReinforcementLB}. Moreover, they improved the learning strategy of probability model to develop a three-phase local search, that is, a starting coloring generation phase based on a probability matrix, a heuristic coloring improvement phase and a learning based probability updating phase~\cite{Zhou2018ImprovingPL}.

\subsection{Population-based metaheuristics for GCPs}

Population-based iteration mechanisms are incorporated to the improve exploration abilities of metaheuristics as well. Hsu \emph{et al.}~\cite{hsu2011mtpso} proposed a modified turbulent particle swarm optimization algorithm for the planar graph coloring problem, where a three-stage turbulent model is employed to strike a balance between  exploration and  exploitation.
Hern{\'a}ndez and Blum~\cite{hernandez2012distributed} dealt with the problem of finding valid graphs colorings in a distributed way, and the assignment of different colors to neighboring nodes is asynchronously implemented by simulating the calling behavior of Japanese tree frogs.
Rebollo-Ruiz and M.~Gra{\~n}a~\cite{rebollo2014empirical} addressed the GCP by a gravitational swarm intelligence algorithm, where nodes of a graph are mapped to agents, and its connectivity is mapped into a repulsive force between the agents corresponding to adjacent nodes.
Based on the conflict matrices of candidate solutions, Zhao \emph{et al.}~\cite{Zhao2020DiscreteSH} developed a dimension-by-dimension update method, by which a discrete selfish herd optimizer was proposed to address GCPs efficiently. Aiming to develop an efficient parameter-free algorithm, Chalupa and Nielsen~\cite{chalupa2021parameter} proposed to improve the global exploration by a multiple cooperative searching strategy. For the four-colormap problem, Zhong \emph{et al.}~\cite{Zhong2022} proposed an enhanced discrete dragonfly algorithm that performs a global search and a local search alternately to color maps efficiently.

The incorporation of probability models are likewise employed to improve the performance of population-based metaheuristics. Bui \emph{et al.}~\cite{Bui2008AnAA} developed a constructive strategy of coloring scheme based on an ant colony, where an ant colors just a portion of the graph unsing only local information. Djelloul \emph{et al.} \cite{Djelloul2015QuantumIC} took a collection of quantum matrices as the population of the cuckoo search algorithm, and an adapted hybrid quantum mutation operation was introduced to get enhanced performance of the cuckoo search algorithm.

\subsection{Hybrid metaheuristics for GCPs}
The population-based metaheuristics can be further improved by hybrid search strategies. Paying particular attention to ensuring the population diversity,
L\"{u} and Hao~\cite{lu2010memetic} proposed  an adaptive multi-parent crossover operator
and a diversity-preserving strategy to improve the searching efficiency of evolutionary algorithms, and proposed a memetic algorithm that takes the TS as a local search engine.
Porumbel \emph{et al.}~\cite{porumbel2010evolutionary} developed a population management strategy that decides whether an offspring should be accepted in the population, which individual needs to be replaced and when mutation is applied.
Mahmoudi and Lotfi~\cite{Mahmoudi2015ModifiedCO} proposed a discrete cuckoo optimization algorithm for the GCP, where a neighborhood search in radius of the lay egg causes the algorithm hardly trapped in local minimum and producing new eggs. Accordingly, it provides a good balance between diversification and centralizing.


Wu and Hao~\cite{wu2012coloring} proposed a preprocessing method that extracts large independent sets by the TS, and the memetic algorithm proposed by L\"{u} and Hao~\cite{lu2010memetic} was employed to color the residual graph.
For the chromatic problem, Douiri and Elbernoussi~\cite{Douiri2015SolvingTG} initialized the color number by the coloring result of a heuristic algorithm and  generated the initial population of genetic algorithm (GA) by finding  a maximal independent  set approximation of the investigated graph.

Bessedik \emph{et al.}~\cite{Bessedik2011HowCB} addressed the GCPs within the framework of the honey bees optimization, where a local search, a tabu search and an ant colony system are implemented as workers and queens are randomly generated. Mirsaleh and Meybodi~\cite{Mirsaleh2018} proposed a Michigan memetic algorithm for GCPs, where each chromosome is associated to a vertex of the input graph. Accordingly, each chromosome is a part of the solution and represents a color for its corresponding vertex, and each chromosome locally evolves by evolutionary operators and improves by a learning automata based local search.
Moalic and Gondran~\cite{moalic2018variations} integrated a TS procedure with an evolutionary algorithm equipped with the greedy partition crossover, by which the hybrid algorithm can performs well with a population consisting of two individuals.
Silva \emph{et al.}~\cite{silva2020improved} developed a hybrid algorithm \emph{iColourAnt}, which addresses the GCP using an ant colony optimization procedure with assistance of a local search performed by the reactive TS.

\subsection{Related work on the estimation of distribution algorithm}


A large number of works have been reported to improve the general performance of EDAs.
To improve the general precision of a distribution model, Shim \emph{et al.}~\cite{Shim2013b} modelled the restricted Boltzmann machine as a novel EDA, where the probabilistic model is constructed using its energy function, and the $k$-means clustering was employed to group the population into small clusters.
Approximating the Boltzmann distribution by a Gaussian model, Valdez \emph{et al.}~\cite{Valdez2013} proposed a Boltzmann univariate marginal distribution algorithm, where the Gaussian distribution obtains a better bias to sample intensively the most promising regions. Considering the multivariate dependencies between continuous random variables, PourMohammadBagher \emph{et al.}~\cite{PourMohammadBagher2017} proposed a parallel model of some subgraphs with a smaller number of variables to avoid complex approximations of learning a probabilistic graphical model.
Dong \emph{et al.}~\cite{Dong2019} proposed a latent space-based EDA, which transforms the multivariate probabilistic model of Gaussian-based EDA into its principal component latent subspace of lower dimensionality to improve its performance on large-scale optimization problems.


To enhance the local exploitation of an EDA, Zhou \emph{et al.}~\cite{Zhou2015} suggested to combine an estimation of distribution algorithm with cheap and expensive local search methods for making use of both global statistical information and individual location information.
Considering that the random sampling of Gaussian EDA usually suffers from the poor diversity and the premature convergence, Dang \emph{et al.}~\cite{Dang2022} developed an efficient mixture sampling model to achieve a good tradeoff between the diversity and the convergence, by which it can explore more promising regions and utilize the unsuccessful mutation vectors.

The performance of EDA can also improved by designing tailored update strategies of the probability model.
To address the multiple global optima of multimodal problem optimizations, P\~{e}na \emph{et al.}~\cite{Pena2005} introduced the unsupervised learning of Bayesian metwokrs in EDA, which makes it able to model simultaneously the different basins represented by the selected individuals, whereas preventing genetic drift as much as possible.
Peng \emph{et al.}~\cite{Peng2015} developed an explicit detection mechanism of the promising areas, by which function evaluations for exploration can be significantly reduced.
To prevent the Gaussian EDAs from premature convergence, Ren \emph{et al.}~\cite{Ren2018} proposed to tune the  main search direction by the anisotropic adaptive variance that is scaled along different eigendirections based on the landscape characteristics captured by a simple topology-based detection method.
Liang \emph{et al.}~\cite{Liang2020} proposed to archive
a certain number of high-quality solutions generated in the previous generations, by which fewer individuals are needed in the current population for model estimation.
In order to address the mixed-variable newsvendor problem, Wang \emph{et al.}~\cite{Wang2020} developed a histogram model-based estimation of distribution algorithm, where an adaptive-width histogram model is used to deal with the continuous variables and a learning-based histogram model is applied to deal with the discrete variables.
Liu \emph{et al.}~\cite{Liu2021} embedded within the search procedure a learning mechanism based on an incremental Gaussian mixture model,  by which all new solutions generated during the evolution are  fed incrementally into the learning model to adaptively discover the structure of the Pareto set of an MOP.

\section{The Distribution Model for the Graph Coloring Problem}\label{DisMod}
\subsection{The graph coloring problem}
Let $n=|V|$ be the vertex number of a graph $G=(V,E)$. An assignment of vertexes with $k$ colors can be represented by an integer vector $\mathbf{x}=(x_1,\dots,x_n)$, where $x_{j}$ denotes the assigned color of vertex $v_{j}$. Then, the $k$-coloring problem can be modelled as a minimization problem
\begin{equation}\label{k-color}
\begin{array}{l}
   \min\quad f_k(\mathbf{x})=\sum\limits_{j_1=1}^n\sum\limits_{j_2=1}^n \delta(j_1,j_2)\\
   s.t.\left\{
  \begin{aligned}
  & \delta(j_1,j_2)=\left\{\begin{aligned}& 1,&&\mbox{if } (v_{j_1}, v_{j_2})\in E \wedge x_{j_1}=x_{j_2},\\ &0, && \mbox{otherwise },\end{aligned}\right.\\
  & \mathbf{x}=(x_1,\dots,x_n), x_{j}\in\{1,\dots,k\}, j=1,\dots,n,\\
  & j_i \in \{1,2,\dots,n\}, i=1,2.
  \end{aligned}
  \right.
\end{array}
\end{equation}

While $\delta(j_1,j_2)=1$, the adjacent vertexes $v_{j_1}$ and $v_{j_2}$ are assigned with the same color, and $(v_{j_1},v_{j_2})$ is called a \emph{conflicting edge}. Accordingly, the objective values $f_k(\mathbf{x})$ represents the total \emph{conflicting number} of the color assignment $\mathbf{x}$.  While $G$ is $k$-colorable, there exists an optimal partition $\mathbf{x}^*$ such that $f_k(\mathbf{x}^*)=0$, and $\mathbf{x}^*$ is called a \emph{legal $k$-color assignment} of graph $G$. Thus, the \emph{chromatic problem} is modelled as
\begin{equation}\label{chro-pro}
\begin{array}{l}
  \min\quad k \\
  s.t.\quad f_k(\mathbf{x}^*)=0,
 \end{array}
\end{equation}
where $\mathbf{x}^*$ represents a legal $k$-color assignment that is an optimal color assignment of problem (\ref{k-color}).

{\color{red}
\subsection{The Q-bit model and the Q-gate transformation}
Different from the ACO and the EDA, the QEA employs a quantum matrix for probabilistic modelling of the solution space, modelling the probability distribution of a binary variable $bx$ by a \emph{Q-bit} $(\alpha,\beta)^T$ satisfying $|\alpha|^2+|\beta|^2=1$~\cite{Han2002QuantuminspiredEA}. That is, $|\alpha|^2$ and $|\beta|^2$ give the probabilities of $bx=1$ and $bx=0$, respectively. Accordingly, the probability distribution of an $n$-dimensional binary vector $\mathbf{bx}=(bx_1,\dots,bx_n)$ is represented by
\begin{equation*}
\mathbf{q}=(\vec{q}_1,\vec{q}_2,\dots,\vec{q}_n)=\left[
\begin{array}{cccc}
  \alpha_1 & \alpha_2 & \cdots & \alpha_n \\
  \beta_1 & \beta_2 & \cdots & \beta_n
\end{array}
\right],
\end{equation*}
where $|\alpha|_j^2+|\beta|_j^2=1$, $j=1,\dots,n$. Then, the value of $bx_j$ can be obtained by sampling the probability distribution $(|\alpha|_j^2,|\beta|_j^2)$.

The probability distribution of binary variable is modified in the QEA by the \emph{Q-gate}, a $2\times 2$ orthogonal matrix
\begin{equation*}
U(\Delta\theta)=\left[
\begin{array}{lr}
  \cos(\Delta\theta)  & -\sin(\Delta\theta) \\
  \sin(\Delta\theta)  & \cos(\Delta\theta)
\end{array}
\right].
\end{equation*}
Premultiplying $\vec{q}_j$ by $U(\Delta\theta)$, the probability distribution of $bx_j$ is modified as
\begin{equation*}
  \vec{q}'_j=U(\Delta\theta)\cdot\vec{q}_j,\quad i=j,\dots,n.
\end{equation*}

To coloring a graph of $n$ vertexes by $k$ colors, Djelloul \emph{et al.} \cite{Djelloul2015QuantumIC} modelled the probability distribution of color assignment by a quantum matrix
\begin{equation*}
\mathbf{q}=(\vec{q}_1,\vec{q}_2,\dots,\vec{q}_n)=\left[
\begin{array}{cccc}
  q_{1,1} & q_{1,2} & \cdots & q_{1,n} \\
  q_{2,1} & q_{2,2} & \cdots & q_{2,n}  \\
  \cdots & \cdots & \cdots & \\
  q_{2k-1,1} & q_{2k-1,2} & \cdots & q_{2k-1,n} \\
  q_{2k,1} & q_{2k,2} & \cdots & q_{2k,n}  \\
\end{array}
\right],
\end{equation*}
where $$(q_{2i-1,j})^2+(q_{2i,j})^2=1,\quad \forall\, i\in\{1,\dots,k\}, j\in\{1,\dots,n\}.$$ In this way, the Q-gate can be deployed for each unit vector $(q_{2i-1,j}, q_{2i,j})^T$ to regulate the distribution of color assignment.
}

\subsection{The proposed distribution model of color assignment}\label{R}
Since the unit vector $(q_{2i-1,j}, q_{2i,j})^T$ models each candidate color $i$ of vertex $j$ independently, the sampling process would lead to multiple assignments for vertex $j$, and an additional repair strategy is needed to get a feasible $k$-coloring assignment \cite{Djelloul2015QuantumIC}.
  To address this defect, we propose to model the color distribution of vertex $j$ by a $k$-dimensional unit vector $\vec{q}_j$, and present the color distribution of $n$ vertexes as
\begin{equation}\label{PM}
\mathbf{q}=(\vec{q}_1,\dots,\vec{q}_n)=\begin{bmatrix}
q_{11} & q_{12} & ... & q_{1n}\\
q_{21} & q_{22} & ... & q_{2n} \\
\vdots & \vdots & \vdots & \vdots \\
q_{k1} & q_{k2} & ... & q_{kn}
\end{bmatrix},
\end{equation}
where $\vec{q}_j$ satisfies
\begin{equation}\label{UV}\|\vec{q}_j\|_2^2={\sum_{i= 1} ^k q_{ij}^2}=1, \quad\forall\,j=1,\dots, n.\end{equation}

The distribution model confirmed by (\ref{PM}) and (\ref{UV}) incorporates the advantages of models in EDAs and QEAs.
\begin{itemize}
  \item An feasible $k$-coloring of $n$ vertexes can be achieved by successively sampling $n$ columns of $\mathbf{q}$.
  \item The update of probability distribution can be implemented by both orthogonal transformations performed on column vectors of $\mathbf{q}$ and direct manipulations of components that do not change the norms of columns vectors~\footnote{Details for the update process are presented in Section \ref{EffSeach}.}.
\end{itemize}



\section{A Distribution Evolutionary Algorithm Based
on a Population of Probability Model}\label{DEAPPM}
{ \color{blue}

The proposed DEA-PPM solves the GCP based on a distribution population and a solution population. The distribution population consists of individuals representing distribution models of graph coloring, which are updated by an orthogonal exploration strategy and an composite exploitation strategy. Meanwhile, an associated solution population is deployed to exploit the solution space by the TS-based local search. Moreover, an iterative vertex removal strategy and a tailored inherited initialization strategy are introduced to accelerate the procedure of $k$-coloring, which in turn  contributes to its high efficiency of addressing the chromatic problem. Thanks to the cooperative interplay between the distribution population and the solution population, the DEA-PPM with small populations is expected to achieve competitive results of GCPs.}

\subsection{The framework of DEA-PPM}\label{framework}
\begin{algorithm}[!htb]
  \caption{The framework of DEA-PPM}\label{Alg_GCP}
  \KwIn{an undirected graph $G=(V,E)$;}
  \KwOut{the color number $k$, the obtained color assignment $\mathbf{x}^*_{G}$;}

  $gen\leftarrow 0$\;
  initialize the color number $k$\;
  \While{termination-condition 1 is not satisfied}
  {
       reduce $G=(V,E)$ to $G'=(V',E')$ by the IVR strategy \;
       \eIf{$gen=0$}
       {initialize $\mathbf{Q}(0)$ by (\ref{ini1});   \\
       sample $\mathbf{Q}(0)$ to generate $\mathbf{P}(0)$; $/*$ \emph{uinform initialization} $*/$
       }
       {
       $(\mathbf{Q}(0),\mathbf{P}(0))=InherInit(\mathbf{Q},\mathbf{P},k)$; $/*$ \emph{inherited initialization} $*/$
       $k=k-1$\;}

       set $\mathbf{x}^*_{G'}$ as the best solution in $\mathbf{P}(0)$\;
       $\mathbf{p}_1=\mathbf{x}^*_{G'}$, $\mathbf{p}_2=\mathbf{x}^*_{G'}$\;

       $t\leftarrow 1$\;

	  \While {termination-condition 2 is not satisfied}
      {
            $\mathbf{Q}'(t)=OrthExpQ(\mathbf{Q}(t-1),\mathbf{P}(t-1))$; $/*$ \emph{orthogonal exploration} $*/$ \\
            $\mathbf{P}'(t)=SampleP(\mathbf{Q}'(t),\mathbf{P}(t-1)$; $/*$ \emph{ sampling with inheritance} $*/$\\
            $(\mathbf{P}(t),\mathbf{p}_1,\mathbf{p}_2,\mathbf{x}^*_{G'})=RefineP(\mathbf{P}'(t),\mathbf{p}_1,\mathbf{p}_2,\mathbf{x}^*_{G'})$ $/*$ \emph{ refinement of the solution population} $*/$\\	
            	
$\mathbf{Q}(t)=RefineQ(\mathbf{P}'(t),\mathbf{P}(t),\mathbf{Q}'(t))$; $/*$ \emph{ refinement of the distribution population} $*/$\\
$t\leftarrow t+1$\;	

      }
      recover $\mathbf{x}^{*}_{G'}$ by the IR strategy to get $\mathbf{x}^{*}_{G}$\;
      $\mathbf{Q}=\mathbf{Q}(t)$, $\mathbf{P}=\mathbf{P}(t)$\;
      $gen\leftarrow gen+1$\;
  }
\end{algorithm}
{\color{red} As presented in Algorithm \ref{Alg_GCP}, DEA-PPM is implemented as two nested loops: the inner loop addressing the $k$-coloring problem  and the outer loop  decreasing $k$ to get the chromatic number $\chi(G)$.}
Based on a distribution population $\mathbf{Q}(t)=(\mathbf{q}^{[1]}(t),\dots, \mathbf{q}^{[np]}(t))$ and the corresponding solution population $\mathbf{P}(t)=(\mathbf{x}^{[1]}(t),\dots, \mathbf{x}^{[np]}(t))$, it starts with the initialization of the color number $k$, which is then minimized by the outer loop to get the chromatic number $\chi(G)$.

Each iteration of the outer loop begins with  the iterative vertex removal (IVR) strategy~\cite{Yu2011LayoutDF}, by which the investigated graph $G$ could be transformed into a reduced graph $G'=(V',E')$, and the complexity of the coloring process could be reduced as well. Then, Lines 5-11 of Algorithm \ref{Alg_GCP} initialize a distribution population $\mathbf{Q}(0)$ and the corresponding solution population $\mathbf{P}(0)$ for $G'$. Once $G'$ is colored by Lines 12-20 of Algorithm \ref{Alg_GCP}, DEA-PPM recovers the obtained color assignment $\mathbf{x}^*_{G'}$ to get an color assignment $\mathbf{x}^*_{G}$ of $G$, and $\mathbf{Q}(t)$ as well as $\mathbf{P}(t)$ is archived for the inherited initialization performed at the next generation. Repeating the aforementioned process until  the \emph{termination-condition 1} is satisfied, DEA-PPM returns a color number $k$ and the corresponding color assignment $\mathbf{x}^*_G$.

After the initialization of $\mathbf{x}^*_{G'}$, $\mathbf{p}_1$, $\mathbf{p}_2$ and $t$, the inner loop tries to get a legal $k$-coloring assignment for the reduced graph $G'$ by evolving both the distribution population $\mathbf{Q}(t)$ and the solution population $\mathbf{P}(t)$. It first performs the orthogonal exploration on $\mathbf{Q}(t)$ to generate $\mathbf{Q}'(t)$, and then, generates an intermediate solution population $\mathbf{P}'(t)$, which is further refined to get $\mathbf{P}(t+1)$. Meanwhile, $\mathbf{Q}(t+1)$ is generated by refining $\mathbf{Q}'(t)$. The inner loop repeats until the \emph{termination-condition 2} is satisfied.

The outer loop of DEA-PPM is implemented  only once for the $k$-coloring problem.  To address the chromatic problem, the \emph{termination condition 1} is satisfied if the chromatic number has been identified or the inner loop fails to get a legal $k$-coloring assignment for a given iteration budget. The \emph{termination condition 2} is met while a legal $k$-coloring assignment is obtained or the maximum iteration number is reached.

For the $k$-coloring problem, DEA-PPM initializes the color number $k$ by a given positive integer.  While it is employed to address the chromatic number problem, we set $k=\Delta G+1$\footnote{Here $\Delta G$ is the maximum vertex degree of graph $G$.} because an undirected graph $G$ is sure to be $(\Delta G+1)$-colorable \cite{Hedetniemi2003LinearTS}.

\subsection{The iterative vertex removal strategy and the inverse recovery strategy}
To reduce the time complexity of the $k$-coloring algorithm,  Yu \emph{et al.}~\cite{Yu2011LayoutDF} proposed an iterative vertex removal (IVR) strategy  to reduce the size of the investigated graph. By successively removing vertexes with degrees less than $k$, IVR generates a reduced graph $G'=(V',E')$, and put the removed vertexes into a stack $S$. In this way, one could get a graph $G'$ where degrees of vertexes are greater than or equal to $k$, and its size could be significantly smaller than that of $G$.

\begin{figure}[!hbt]
\centering
\subfigure[]{
\includegraphics[width=3.6cm]{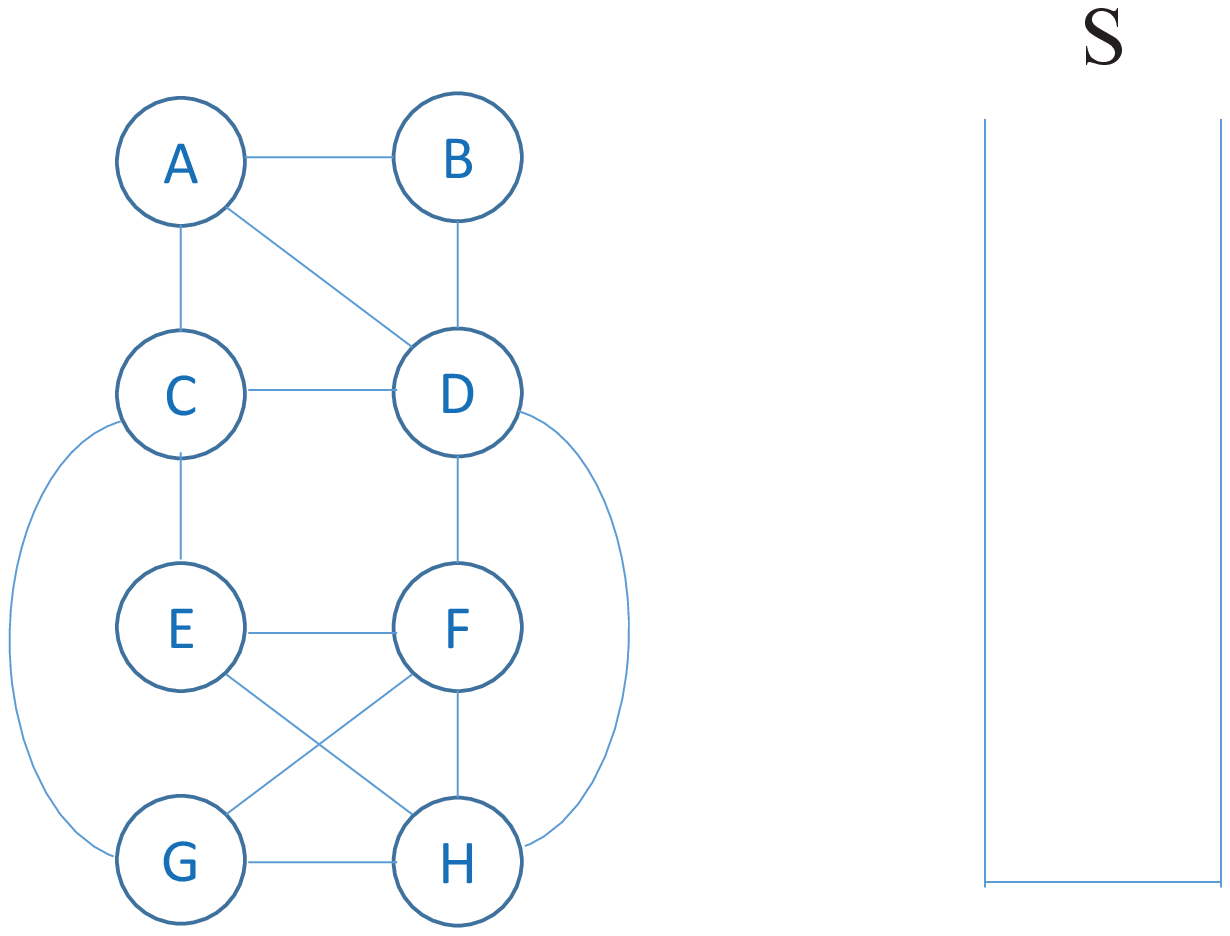}
}\label{IVRFig-1}
\subfigure[ ]{
\includegraphics[width=3.6cm]{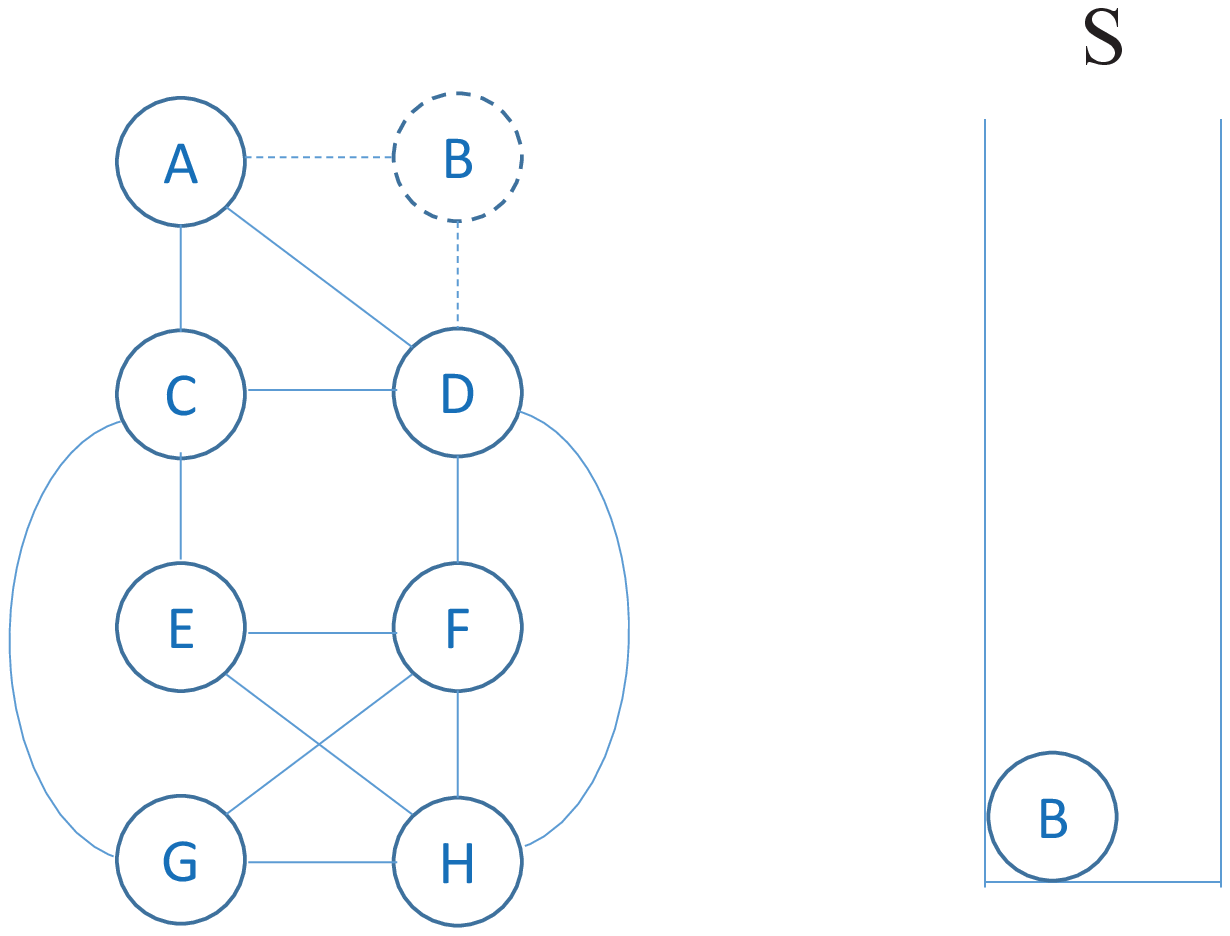}
}\label{IVRFig-2}
\subfigure[ ]{
\includegraphics[width=3.6cm]{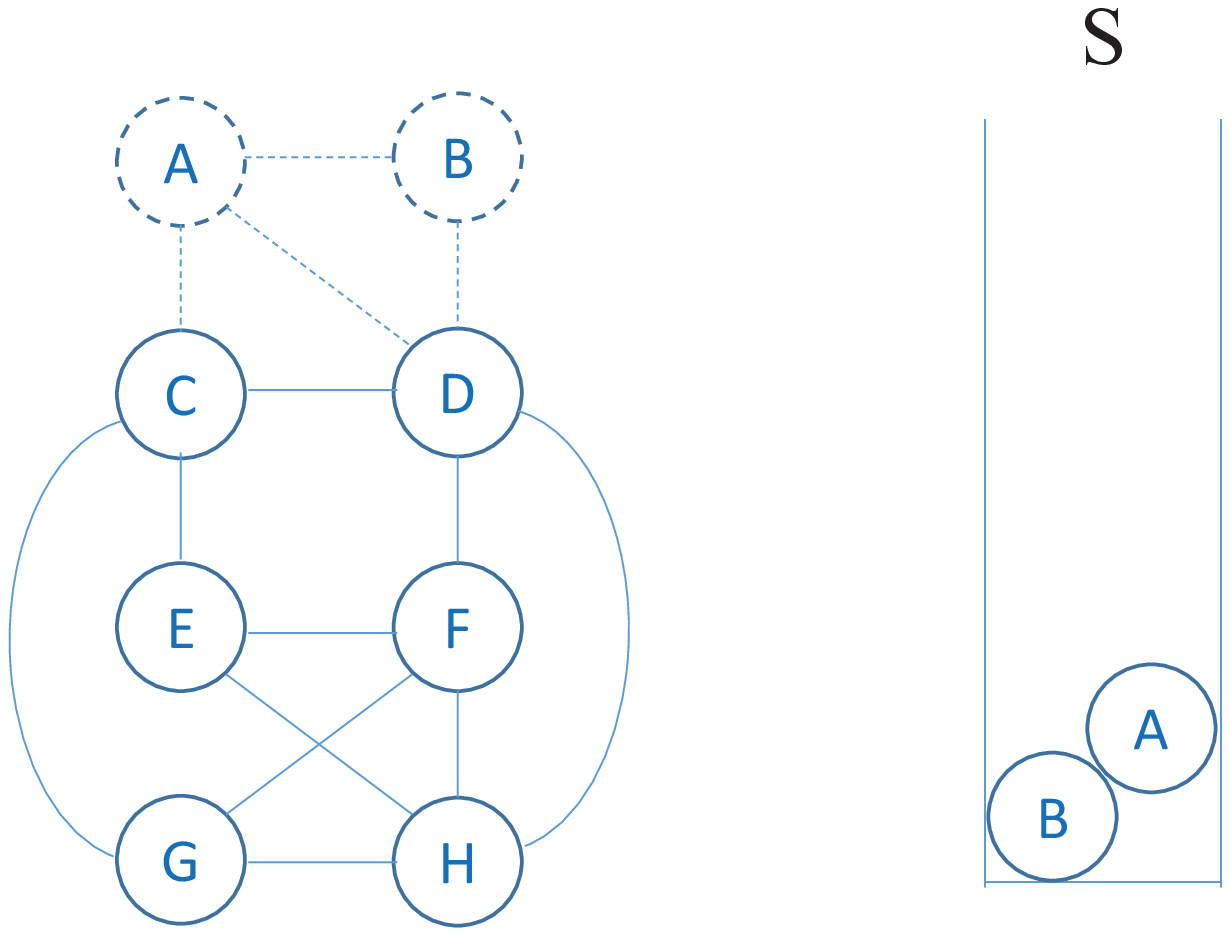}
}\label{IVRFig-3}
\\
\subfigure[ ]{
\includegraphics[width=3.6cm]{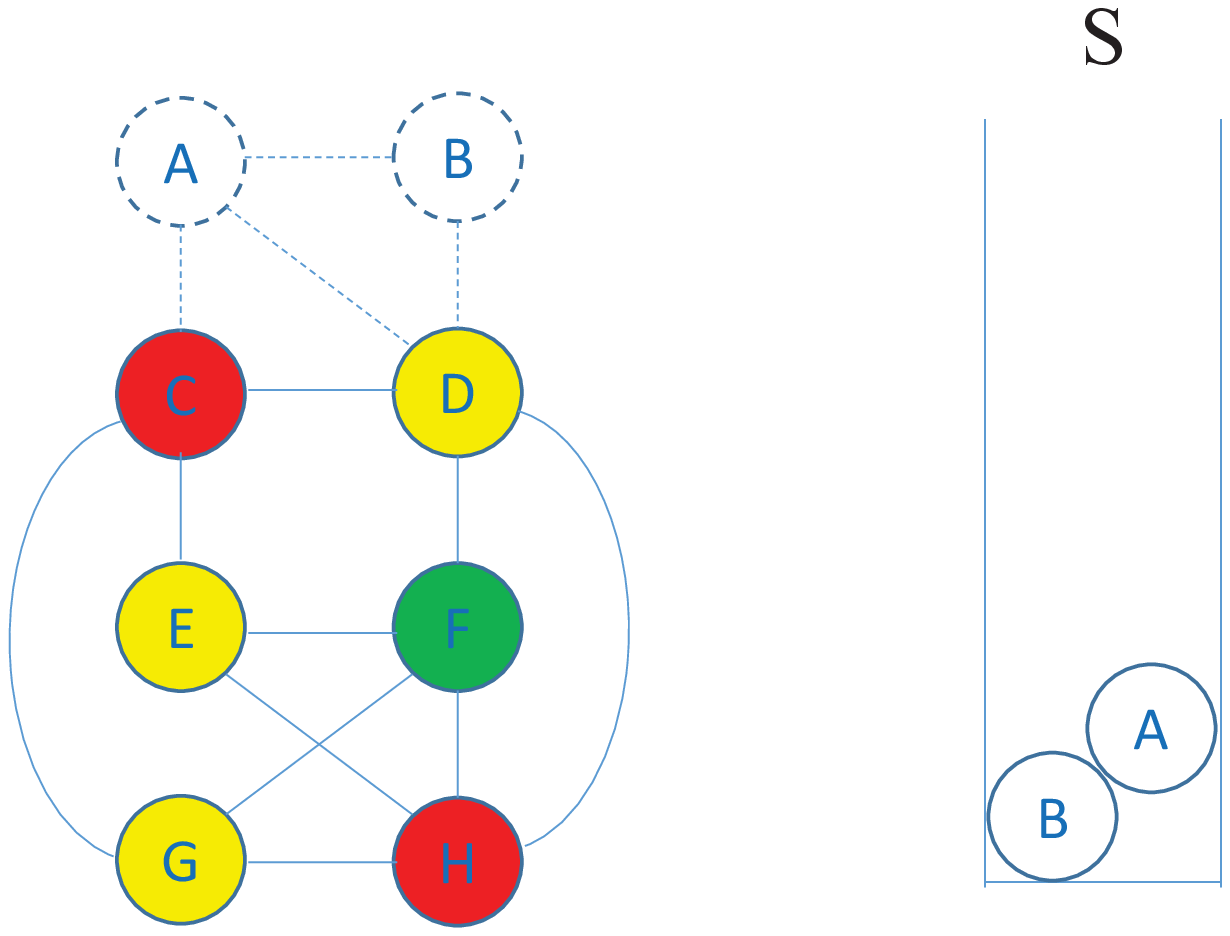}
}\label{IVRFig-4}
\subfigure[ ]{
\includegraphics[width=3.6cm]{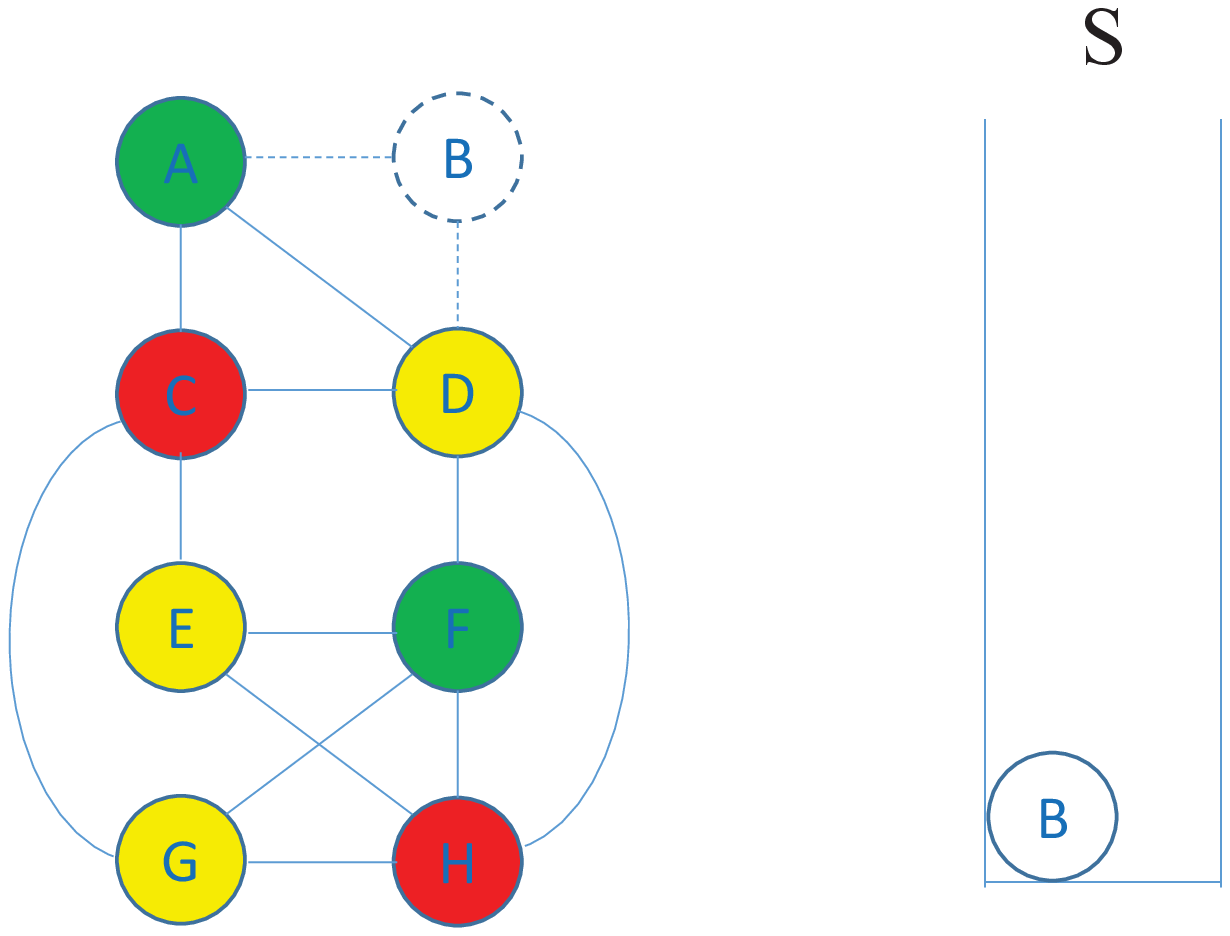}
}\label{IVRFig-5}
\subfigure[ ]{
\includegraphics[width=3.6cm]{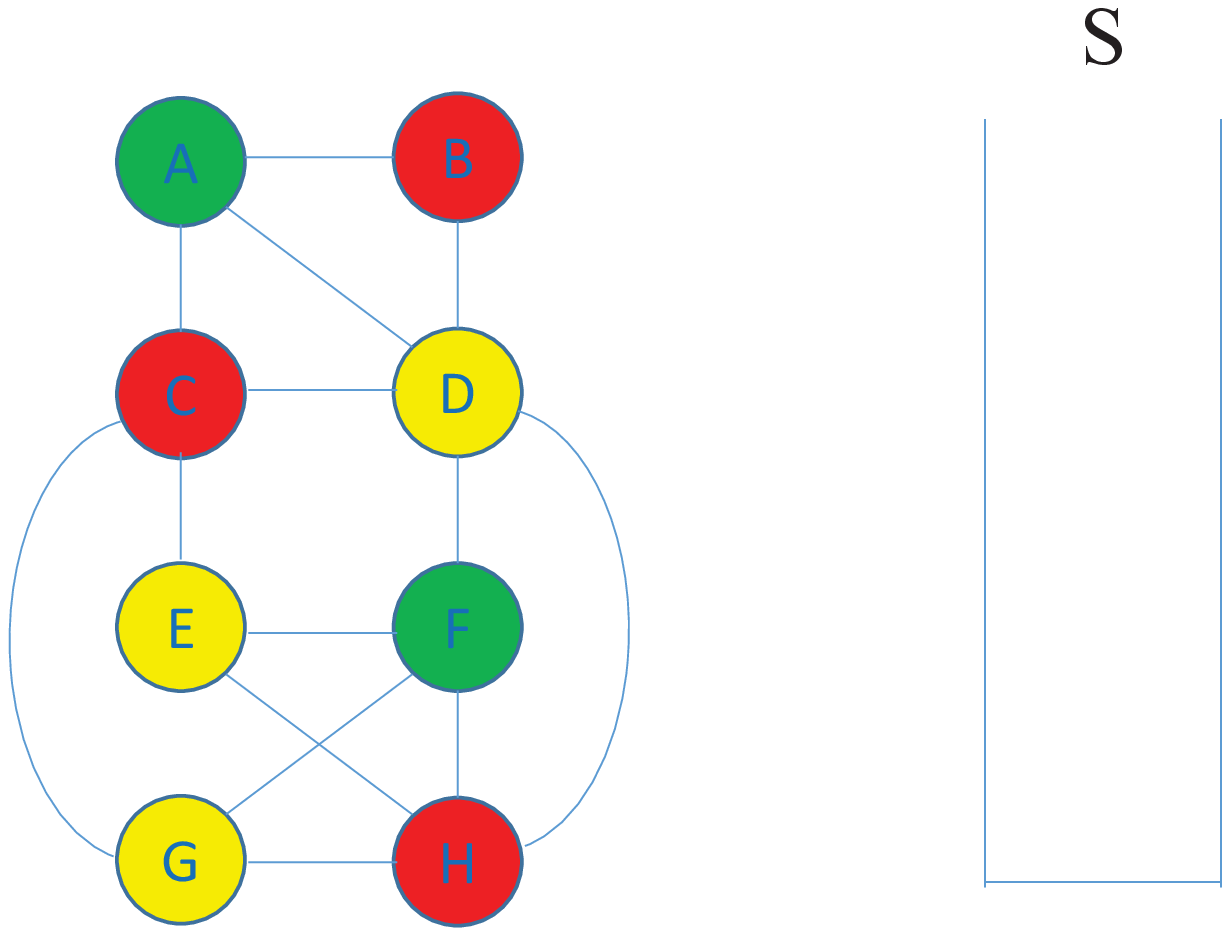}
}\label{IVRFig-6}
\caption{The iterative vertex removal and the inverse recovery process for the 3-coloring of $G=(V,E)$. (a) Graph $G=(V,E)$ consisting of 8 vertexes and 14 edges. (b)(c) Vertexes $B$ and $A$ are  successively removed from $G$ to get the reduced graph $G'=(V',E')$. (d) A legal 3-color assignment of $G'$. (e)(f) Recover vertexes $A$ and $B$ to get a legal 3-color assignment of $G$.}
\label{IVRFig}
\end{figure}

While a $k$-coloring assignment $\mathbf{x}^{*}_{G'}$ is obtained for the reduced graph $G'$, the inverse recovery (IR) operation is implemented by recovering vertexes in the stack $S$. The IR process is initialized by assigning any legal color to the vertex at the top of  $S$. Because the IVR process removes vertexes with degree less than $k$, the IR process  can get all recovered vertexes colored without conflicting. An illustration for the implement of the IVR and the IR is presented in Fig. \ref{IVRFig}.

\subsection{Population initialization}

Depending on the iteration stage of DEA-PPM, the initialization of populations is implemented  by the \emph{uniform initialization} or the \emph{inherited initialization}.

At the beginning, the uniform initialization generates $np$ individuals of $\mathbf{Q}(0)$  as
\begin{equation}\label{ini1}
\mathbf{q}^{(0)} =
\begin{bmatrix}
\frac{1}{\sqrt{k}} & \frac{1}{\sqrt{k}} & ... &\frac{1}{\sqrt{k}}\\
\frac{1}{\sqrt{k}} & \frac{1}{\sqrt{k}} & ... & \frac{1}{\sqrt{k}} \\
\vdots & \vdots & \vdots & \vdots \\
\frac{1}{\sqrt{k}} & \frac{1}{\sqrt{k}} & ... & \frac{1}{\sqrt{k}}
\end{bmatrix}.
\end{equation}
$\mathbf{P}(0)$ are generated by sampling model (\ref{ini1}) $np$ times.

While $gen >0$, the \emph{inherited initialization} gets $\mathbf{Q}(0)$ and $\mathbf{P}(0)$ with the assistance of the distribution populations $\mathbf{Q}$ and the solution population $\mathbf{P}$ archived at the last generation. The graph $G$ has been colored with $k$ colors, and it is anticipated to get a legal $k-1$-color assignment. To get an initial color assignment of $k-1$ colors, a color index $l_m$ that corresponds to the minimum vertex independent set is identified for $\mathbf{x}^{[i]}=({x}^{[i]}_1,\dots,{x}^{[i]}_n)\in\mathbf{P}$. Then, we get the initial color assignment $\mathbf{y}^{[i]}=({y}^{[i]}_1,\dots,{y}^{[i]}_n)$ by
\begin{equation*}
  y^{[i]}= \begin{cases}
             x^{[i]}-1, & \mbox{if } x^{[i]}\ge l_m, \\
             x^{[i]}, & \mbox{otherwise.}
           \end{cases}
\end{equation*}
Meanwhile, delete the $l_m$-th row of $\mathbf{q}^{[i]}$, and normalize its columns to get an initial distribution $\mathbf{r}^{[i]}$ corresponding to $k-1$ colors.
Details of the \emph{inherited initialization} are presented in Algorithm \ref{AlgInit}.

%
%
%

\begin{algorithm}[!htb]
  \caption{$(\mathbf{Q}',\mathbf{P}')=InherInit(\mathbf{Q},\mathbf{P},k)$}\label{AlgInit}
  \KwIn{a distribution population $\mathbf{Q}$, a solution population $\mathbf{P}$, a color number $k$;}
  \KwOut{the initialized distribution population $\mathbf{Q}'$, the initialized solution population $\mathbf{P}'$;}
  \For{$i=1,\dots,np$}
  {$\mathbf{x}^{[i]}\in \mathbf{P}$,$\mathbf{q}^{[i]}\in \mathbf{Q}$\;
  transform $\mathbf{x}^{[i]}=(x^{[i]}_1,\dots,x^{[i]}_n)$ to a vertex partition $s=\{V_{1},\dots,V_{k}\}$\;
  $l_m\leftarrow \arg\min\{|V_j|\}$\;
  \For{$j=1,\dots,n$}
  {
  \If{$x^{[i]}_j\ge l_m$}
  {$x^{[i]}_j=x^{[i]}_j-1$\;}
  }
  $\mathbf{y}^{[i]}=\mathbf{x}^{[i]}$\;
  delete the $l_m$-th row of $\mathbf{q}^{[i]}$ and  normalize $\mathbf{q}^{[i]}$ to get $\mathbf{r}^{[i]}$\;
  }
  $\mathbf{Q}'=\bigcup_{i=1}^{np}\mathbf{r}^{[i]}$, $\mathbf{P}'=\bigcup_{i=1}^{np}\mathbf{y}^{[i]}$\;
\end{algorithm}

\subsection{Evolution of the distribution population}\label{EffSeach}
Based on the distribution model defined by (\ref{PM}) and (\ref{UV}), DEA-PPM performs the \emph{orthogonal exploration} on individuals of $\mathbf{Q}(t)$ to explore the probability space. Moreover, distribution individuals of $\mathbf{Q}(t)$ are refined by  an \emph{exploitation strategy} or a \emph{disturbance strategy}.

\subsubsection{Orthogonal transformation}
An orthogonal transformation on a column vector is performed by premultiplying an orthogonal matrix ${M}$, a square matrix satisfying
$$M^T\cdot M=M\cdot M^T=I,$$ where $ I$ is the identity matrix. Because an orthogonal transformation preserves the 2-norm~\cite{greub1975linear}, we know
\begin{equation}
\label{PN}\|M\vec{v}\|_2=\|{\vec{v}}\|_2,\quad\forall \vec{v}\,\in\mathbb{R}^n.
\end{equation}
Then, by performing orthogonal transformations on columns of the distribution individuals, DEA-PPM can explore the distribution space flexibly.

Since columns of an orthogonal matrix are orthonormal~\cite{greub1975linear}, one can get an orthogonal matrix by performing the QR decomposition on an invertible matrix~\cite{kress1998numrical}.

\subsubsection{Orthogonal exploration in the distribution space}\label{OE}
To perform the orthogonal exploration in the distribution space, DEA-PPM generates an orthogonal matrix by performing the QR decomposition on an invertible matrix that is generated randomly. As presented in Algorithm \ref{AlgOrth}, $m$ worst individuals of $\mathbf{Q}$ are modified by random orthogonal transformations performed on $c$ randomly selected columns. As an initial study, $m$ is set as a random integer in $[1,np/2]$, and $c$ is an integer randomly sampled in $[1,n/10]$.

\begin{algorithm}[!hbt]
    \caption{$\mathbf{Q'}=OrthExpQ(\mathbf{Q},\mathbf{P})$}\label{AlgOrth}
    \KwIn{a distribution population $\mathbf{Q}$, a solution population $\mathbf{P}$;}
  \KwOut{the updated distribution population $\mathbf{Q}'$;}
    sorting  $\mathbf Q$ by fitness values of corresponding individuals in $\mathbf P$\;
    take $\mathbf{Q}_w$ as the collection of $m$ worst individuals of $\mathbf{Q}$\;
    $\mathbf{Q}'=\mathbf{Q}\setminus \mathbf{Q}_w$\;
    \For {$\mathbf{q}\in\mathbf{Q}_w$}
    {
        $\mathbf{q}'\leftarrow\mathbf{q}$\;
        randomly select $c$ columns $\vec{q}'_{j_l} (l=1,...,c)$ from $\mathbf{q}'$\;

        \For{$l=1,...,c$}
        {
        generate a random orthogonal matrix $M_l$ \;
        $\vec{q_{j_l}}'=M_l\vec{q_{j_l}}'$\;
        }
        $\mathbf{Q}'=\mathbf{Q}'\cup \mathbf{q}'$;
    }

\end{algorithm}

\subsubsection{Refinement of the distribution population}\label{update}

\begin{algorithm}[!hbt]
\caption{$\mathbf Q=RefineQ(\mathbf{P}',\mathbf P,\mathbf{Q}')$}\label{Alg_UpQ}
      \KwIn{two solution populations $\mathbf P$ and $\mathbf P'$, a distribution population $\mathbf Q$;}
      \KwOut{the refined distribution population $\mathbf Q'$;}
      \For {$i=1,\dots, np$}
     {
     $\mathbf{q}^{[i]}\in\mathbf{Q}'$, $\mathbf{x}^{[i]}\in\mathbf{P}'$,$\mathbf{y}^{[i]}\in\mathbf{P}$\;
      \For {$j=1,\dots,n$}
      {
        set $rnd_j \sim U(0,1)$\;
        \eIf{$rnd_j\le p_0$}
        {$\vec{r}^{[i]}_j$ is generated by the {exploitation strategy} confirmed by Eqs. (\ref{update11}) and (\ref{update12})\;}
        {$\vec{r}^{[i]}_j$ is generated by  the {exploitation strategy} defined by Eq. (\ref{update14})\;}
      }
$\mathbf{r}^{[i]}=(\vec{r_1}^{[i]},\dots,\vec{r_n}^{[i]})$\;
      }
$\mathbf{Q}=\bigcup_{i=1}^{np}\mathbf{r}^{[i]}$.
\end{algorithm}

As presented in Algorithm \ref{Alg_UpQ}, the distribution population $\mathbf{Q}'$ is refined to generate $\mathbf{Q}$.  $\forall\, \mathbf{q}^{[i]}=({q}^{[i]}_{i,j})_{k\times n}\in\mathbf{Q}'$, DEA-PPM refines its $j$-th column $\vec{q}^{[i]}_j$ with the assistance of the $j$-th components of $\mathbf{x}^{[i]}=(x^{[i]}_1,\dots,x^{[i]}_n)\in \mathbf{P}'$ and $\mathbf{y}^{[i]}=(y^{[i]}_1,\dots,y^{[i]}_n)\in\mathbf{P}$. With probability $p_0$,  $\vec{q}^{[i]}_j$ is refined by an \emph{exploitation strategy}; otherwise, its refinement is implemented by a \emph{disturbance strategy}.


\paragraph{The exploitation strategy} Similar to the probability learning procedure proposed in \cite{Zhou2018ImprovingPL}, the first phase of the exploitation strategy is implemented by
\begin{equation}\label{update11}
  r^{[i]}_{l,j}=\begin{cases}
\sqrt{\alpha +(1-\alpha)(q^{[i]}_{l,j})^{2}}  & \text{ if } l=y^{[i]}_j, \\
 \sqrt{(1-\alpha)(q^{[i]}_{l,j})^{2}} & \text{ if } l\neq y^{[i]}_j,
\end{cases}\quad l=1,\dots,k, 
\end{equation}
where $y^{[i]}_j$ is the $j$-th component of $\mathbf{y}^{[i]}$. Then, an local orthogonal transformation is performed as
\begin{equation}\label{update12}
  \begin{bmatrix}
 r^{[i]}_{l_1,j} \\
 r^{[i]}_{l_2,j}
\end{bmatrix}
=U(\Delta \theta_j )\times \begin{bmatrix}
 r^{[i]}_{l_1,j} \\
 r^{[i]}_{l_2,j}
\end{bmatrix},
\end{equation}
where
\begin{equation*}
  U(\Delta \theta_j )=\begin{bmatrix}
  cos(\Delta \theta_j)& -sin(\Delta \theta_j)\\
  sin(\Delta \theta_j)&cos(\Delta \theta_j)
\end{bmatrix},
\end{equation*}
$l_1=x^{[i]}_j$, $l_2=y^{[i]}_j$.
Equation (\ref{update11}) conducts an overall regulation controlled by the parameter $\alpha$, and equation (\ref{update12})  rotates the subvector $(r^{[i]}_{l_1,j},r^{[i]}_{l_2,j})^T$ counterclockwise by $\Delta \theta_i$ to regulate it slightly.
\paragraph{The disturbance strategy} $\forall\, j\in\{1,2,\dots,n\}$, $\vec{r}^{[i]}_j=(r^{[i]}_{1,j},\dots, r^{[i]}_{k,j})$ is generated by
\begin{equation}\label{update14}
r^{[i]}_{l,j}=  \begin{cases}
  \sqrt{\frac{\lambda (q^{[i]}_{l_0,j})^{2}}{1-(1-\lambda)(q^{[i]}_{l_0,j})^{2} }} & \text{ if } l=l_0, \\
  \sqrt{\frac{(q^{[i]}_{l,j})^{2}}{1-(1-\lambda)(q^{[i]}_{l_0,j})^{2} }} & \text{ if } l\neq l_0,
\end{cases}
\end{equation}
$l=1,\dots,k$. For $0<\lambda <1$, the $l_0$-th components of $\vec{r}^{[i]}_j$ is smaller than that of $\vec{q}^{[i]}_j$, and others are greater. Thus, we set $l_0=y^{[i]}_j$ to prevent DEA-PPM from premature convergence.

\subsection{Efficient search in the solution space}
To search the solution space efficiently, DEA-PPM generates a solution population by \emph{sampling with inheritance}, and then, refines it using a multi-parent crossover operation followed by the TS search proposed in Ref. \cite{Galinier1999HybridEA}.


\subsubsection{The strategy of sampling with inheritance}\label{QMS}

Inspired by the group selection strategy~\cite{Zhou2016ReinforcementLB}, components of new solution $\mathbf{y}^{[i]}=({y}_1^{[i]},\dots,{y}_n^{[i]})$ are either generated by sampling the distribution $\mathbf{q}^{[i]}=(\vec{q}^{[i]}_1,\dots,\vec{q}^{[i]}_n)$ or inheriting from the corresponding solution $\mathbf{x}^{[i]}=({x}_1^{[i]},\dots,{x}_n^{[i]})$. The strategy of sampling with inheritance is presented in Algorithm \ref{Alg_Gen}, where $r$ is the probability of generating $y^{[i]}_j$ by sampling $\vec{q}_j^{[i]}$.
\begin{algorithm}[!hbt]
      \caption{$\mathbf P'=SampleP(\mathbf Q,\mathbf P)$}\label{Alg_Gen}
      \KwIn{a distribution population $\mathbf Q$, a solution population $\mathbf P$;}
      \KwOut{the generated solution population $\mathbf P'$;}
      \For {$i=1,\dots, np$}
     {
     $\mathbf{q}^{[i]}\in\mathbf{Q}$, $\mathbf{x}^{[i]}\in\mathbf{P}$\;
      \For {$j=1,\dots,n$}
      {
        set $rnd_j\sim U(0,1)$\;
        \eIf {$rnd_j<r$}
        {sampling $\vec{q}^{[i]}_j$ to get ${y^{[i]}_j}$\;}
        {${y}^{[i]}_j={x}^{[i]}_j$\;}
      }
      $\mathbf{y}^{[i]}=({y}^{[i]}_1,\dots,{y}^{[i]}_n)$\;

      }
      $\mathbf{P}'=\bigcup_{i=1}^{np}\mathbf{y}^{[i]}$.
\end{algorithm}

\subsubsection{Refinement of the solution population}\label{ls}
The quality of generated solutions is further improved by a refinement strategy presented in Algorithm \ref{Alg_Ref}, which is an iterative process consisting of a multi-parent greedy partition crossover guided by two promising solutions $\mathbf{p}_1$ and $\mathbf{p}_2$ as well as the TS process presented in Ref. \cite{Galinier1999HybridEA}. Meanwhile, two promising solutions $\mathbf{p}_1$ and $\mathbf{p}_2$ are updated.  The refinement process ceases once it stagnates for 20 consecutive iterations.

\begin{algorithm}[!htb]
\caption{$(\mathbf{P},\mathbf{p}_1,\mathbf{p}_2,\mathbf{x}^*_{G'})=RefineP(\mathbf{P}',\mathbf{p}_1,\mathbf{p}_2,\mathbf{x}^*_{G'})$}\label{Alg_Ref}

      \KwIn{a solution population $P'$, two reference solutions $\mathbf{p}_1$ and $\mathbf{p}_2$, the best color assignment $\mathbf{x}^*_{G'}$;}
      \KwOut{the updated solution population $P$, two updated reference solutions $\mathbf{p}_1$ and $\mathbf{p}_2$, the updated best color assignment $\mathbf{x}^*_{G'}$;}
      $iter\leftarrow 0$, $iter\_stag\leftarrow 0$\;
      $\mathbf{c}_1=\mathbf{x}^*_{G'}$\;

      \While {$iter\_stag<20$}
    {
      $\mathbf{P}=MGPX(\mathbf{P}',\mathbf{p}_1,\mathbf{p}_2)$\;
      $\mathbf{P}=Tabu(\mathbf{P})$\;
      record the best solution in $\mathbf{P}$ as $\mathbf{b}$\;
      \eIf{$f(\mathbf{b})<f(\mathbf{p}_1)$}
      { $iter\_stag=0$\;
      $\mathbf{c}_1=\mathbf{p}_1,\mathbf{p}_1=\mathbf{b}$\;}
      {$iter\_stag=iter\_stag+1$\;}

      \If{$f(\mathbf{b})<f(\mathbf{x}^*_{G'})$}
      {$\mathbf{x}^*_{G'}=\mathbf{b}$\;
      }

      \If{$mod(iter,10)=0$}
      {$\mathbf{p}_2=\mathbf{c}_1$\;}
      $\mathbf{P}'=\mathbf{P}$\;
      $iter=iter+1$\;
      }
\end{algorithm}

\paragraph{Multi-parent greedy partition crossover (MGPX)} Inspired by the motivation of greedy partition crossover (GPX) for graph coloring~\cite{Galinier1999HybridEA}, we propose the multi-parent greedy partition crossover (MGPX) presented in Algorithm \ref{Alg_MGPX}. For $\mathbf{x}_1\in \mathbf{P}$, two mutually different solutions $\mathbf{x}_2$ and $\mathbf{x}_3$ are selected from $\mathbf{P}\setminus \{\mathbf{x}_1\}\cup\{\mathbf{p}_1,\mathbf{p}_2\}$. Then, the MGPX is performed on $\mathbf{x}_1$, $\mathbf{x}_2$ and $\mathbf{x}_3$ to generate a new solution $\mathbf{y}$. After the traversal of the solution population $\mathbf{P}$, all generated solutions construct the intermediate solution population $\mathbf{P}'$.

\paragraph{Update of the promising solutions} The promising solution $\mathbf p_1$ is updated if a better solution $\mathbf{b}$ is obtained. Then, $\mathbf p_2$ is set as the original values of $\mathbf p_1$. To fully exploits the promising information incorporated by $\mathbf p_2$, it is updated once every 10 iterations.

\begin{algorithm}[!htb]
      \caption{$\mathbf{P}'=MGPX(\mathbf{P},\mathbf{p}_1,\mathbf{p}_2)$}\label{Alg_MGPX}
      \KwIn{a solution population $\mathbf{P}$, two reference solutions $\mathbf{p}_1$ and $\mathbf{p}_2$;}
      \KwOut{the updated solution population $\mathbf{P}'$;}
$\mathbf{P}'=\emptyset$\;
\For{$\mathbf{x}\in\mathbf{P}$}
      {
      let $\mathbf{x}_1=\mathbf{x}$, $\mathbf{x}_2$ and $\mathbf{x}_3$ be two different solutions selected from $\mathbf{P}\setminus\{\mathbf{x}\}\cup\{\mathbf p_1,\mathbf{p_2}\}$\;
      transform $\mathbf{x}_i$ to the corresponding vertex partition $s_i=\{V_{1}^i,\dots,V_{k}^i\}$, $i=1,2,3$\;
      \For {$l=1,\dots,k$}
    {
      randomly select an index $i_0\in\{1,2,3\}$ according to the probability distribution $\{P_{c},(1-P_{c})/2,(1-P_{c})/2\}$\;
      choose $l_0$ such that $V_{l_0}^{i_0}$ has a maximum cardinality\;
      $V_l:=V_{l_0}^{i_0}$\;
      remove the vertices of $V_l$ from $s_1$, $s_2$ and $s_3$\;
      }
      assign the vertices of $V'\setminus \bigcup_{l=1}^kV_l$ by $s_1$\;
      transform $s=(V_1,\dots , V_k )$ to a solution $\mathbf{y}$\;
      $\mathbf{P}'=\mathbf{P}'\cup \{\mathbf{y}\}$\;
      }
\end{algorithm}

\section{Numerical Experiments}\label{result}

{\color{red} The investigated algorithms are evaluated on the benchmark instances from the second DIMACS competition\footnote{Publicly available at ftp://dimacs.rutgers.edu/pub/challenge/graph/benchmarks/color/.} that were used to test graph coloring algorithms in recent studies~\cite{Galn2017SimpleDG,Sun2021ASM,Zhou2018ImprovingPL,lu2010memetic,moalic2018variations}
.} All tested algorithms are developed in C++ programming language, and run in Microsoft Windows 7 on a laptop equipped with the Intel(R) Core(TM) i7 CPU 860 @ 2.80GHz and 8GB system memory. We first perform a parameter study to get appropriate parameter settings of DEA-PPM, and then, the proposed evolution strategies of distribution population are investigated to demonstrate their impacts on its efficiency. Finally, numerical comparisons for both the chromatic problem and the $k$-coloring problem are performed with the state-of-the-art algorithms. For numerical experiments, time budgets of all algorithms are consistently set as 3600 seconds (one hour). {\color{red} Because performance of the investigated algorithms varies for the selected benchmark problems, we perform the numerical comparison in two different ways. If two compared algorithms achieve inconsistent coloring results for the chromatic problem, numerical comparison is performed by the obtained color numbers; otherwise, we take the running time as the evaluation metric while they get the same coloring results.}

\subsection{Parameter study}\label{parstu}

By setting $k$ as the best known color numbers of the benchmark problems, preliminary experiments for the $k$-coloring problem show  that the performance of DEA-PPM is significantly influenced by the population size $np$, the regulation parameter $\alpha$ and the maximum iteration budget $Iter_{max}$ of the TS. Then, we first demonstrate the univariate influence of parameters by the one-way analysis of variance (ANOVA), and then, perform a descriptive comparison to get a set of parameter for further numerical investigations. {\color{blue} The benchmark instances selected for the parameter study are the $k$-coloring problems of \emph{DSJC500.5}, \emph{flat300\_28\_0}, \emph{flat1000\_50\_0}, \emph{flat1000\_76\_0}, \emph{le450\_15c}, \emph{le450\_15d}}.

{\color{red}
\subsubsection{Analysis of variance on the impacts of parameters}

Our preliminary experiments show that DEA-PPM achieves promising results with $np=8$, $\alpha=0.2$ and $iter_{max}=5000$, which is taken as the baseline parameter setting of the one-side ANOVA test of running time.  With the significance level of 0.05, the significant influences  are highlighted in Tab.\ref{para_anova} by bold P-values.

\begin{table}[!htb]
  \centering
  \caption{Results of the one-way ANOVA test.}\label{para_anova}
\resizebox{\textwidth}{!}
{
  \begin{tabular}{c|c|c|c|c|c|c|c}
    \hline\hline
    \multicolumn{2}{c|}{Tested Parameter} & \multicolumn{6}{c}{P Values}\\
    \hline
    Parameter & Settings   & flat300\_28\_0 & le450\_15c & le450\_15d & DSJC500.5 & flat1000\_50\_0 & flat1000\_76\_0\\
    \hline
    $np$ & $\{4,6,8,10,12\}$  & 0.162 & 0.548 & 0.404 & \textbf{0.004} & \textbf{0.001} & \textbf{0.001}\\
    $\alpha$ & $\{0.1,0.15,0.2,0.25,0.3\}$  & 0.622 & 0.643 & 0.357 & \textbf{0.003} & 0.268 & \textbf{0.000}\\
    $iter_{max}$ & $\{0.5,1,1.5,2.0,2.5\}\times10^4$  & 0.025 & 0.146 & \textbf{0.000} & \textbf{0.040} & \textbf{0.000} & \textbf{0.000}\\
    \hline\hline
  \end{tabular}
  }
\end{table}

Generally, the univariate changes of $np$, $\alpha$ and $iter_{max}$ do not have significant influence on performance of DEA-PPM for instances \emph{flat300\_28\_0}, \emph{le450\_15c} and \emph{le450\_15d}, except that values of $iter_{max}$ has great impact on the results of \emph{le450\_15d}. But for instances \emph{DSJC500.5}, \emph{flat1000\_50\_0} and \emph{flat1000\_76\_0}, the influence is significant, except that $\alpha$ does not significantly influence the performance of DEA-PPM on \emph{flat1000\_50\_0}. To illustrate the results, we included the curves of expected running time in Fig. \ref{fig-para}. The univariate analysis shows that the best results could be achieved by setting $np=8$, $\alpha=0.2$ and $iter_{max}=5000$.

\begin{figure}[!hbt]
\centering
\subfigure[$np$]{
\includegraphics[width=3.9cm]{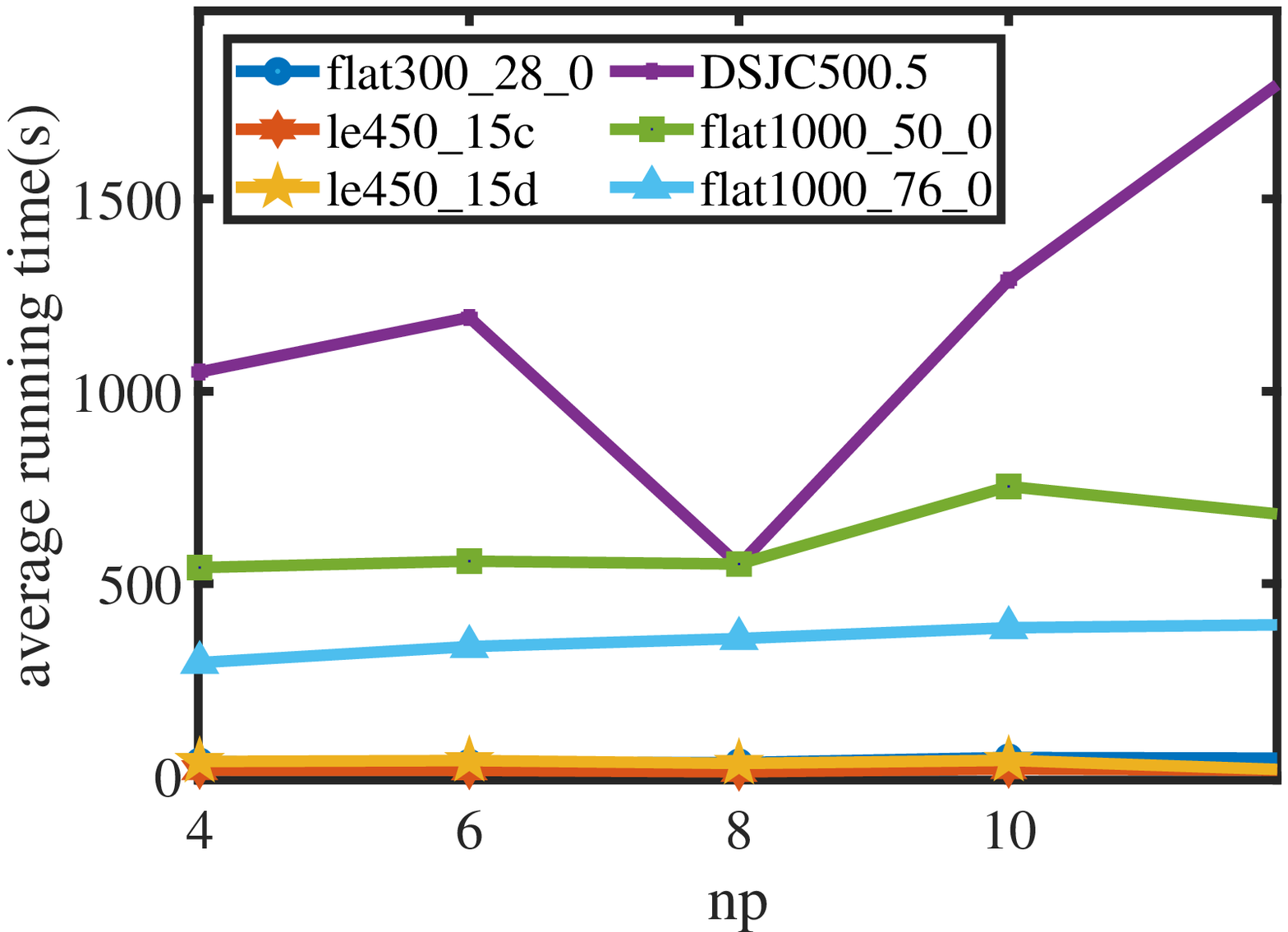}
}\label{para-1}
\subfigure[$\alpha$ ]{
\includegraphics[width=3.4cm]{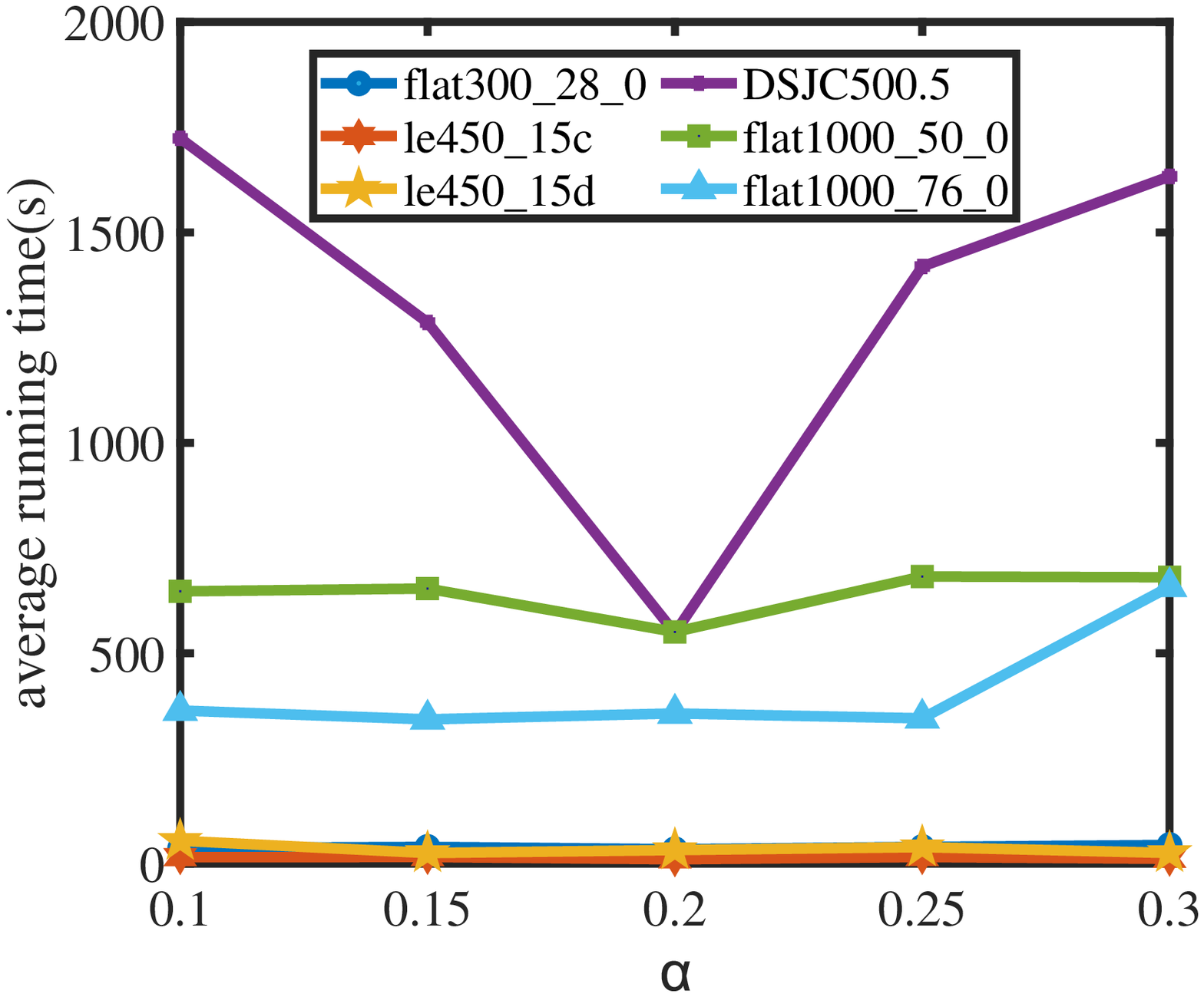}
}\label{para-2}
\subfigure[$iter_{max}$]{
\includegraphics[width=3.9cm]{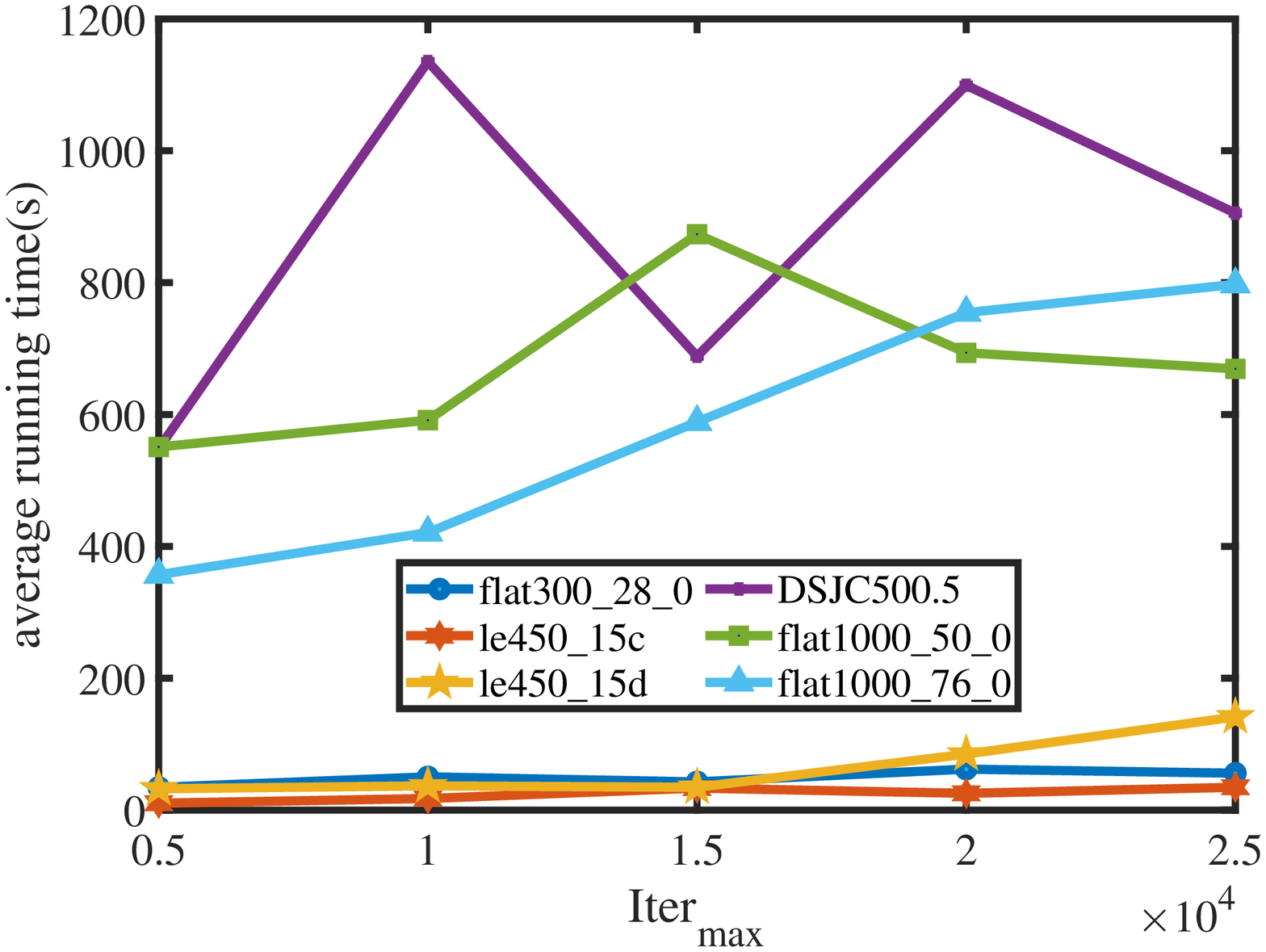}
}\label{para-3}
\caption{Influence of parameters on expected running time of DEA-PPM.}
\label{fig-para}
\end{figure}
}
\subsubsection{Descriptive statistics on the composite impacts of parameters}
\begin{figure}[!hbt]
\centering
\subfigure[]{
\includegraphics[width=5.7cm]{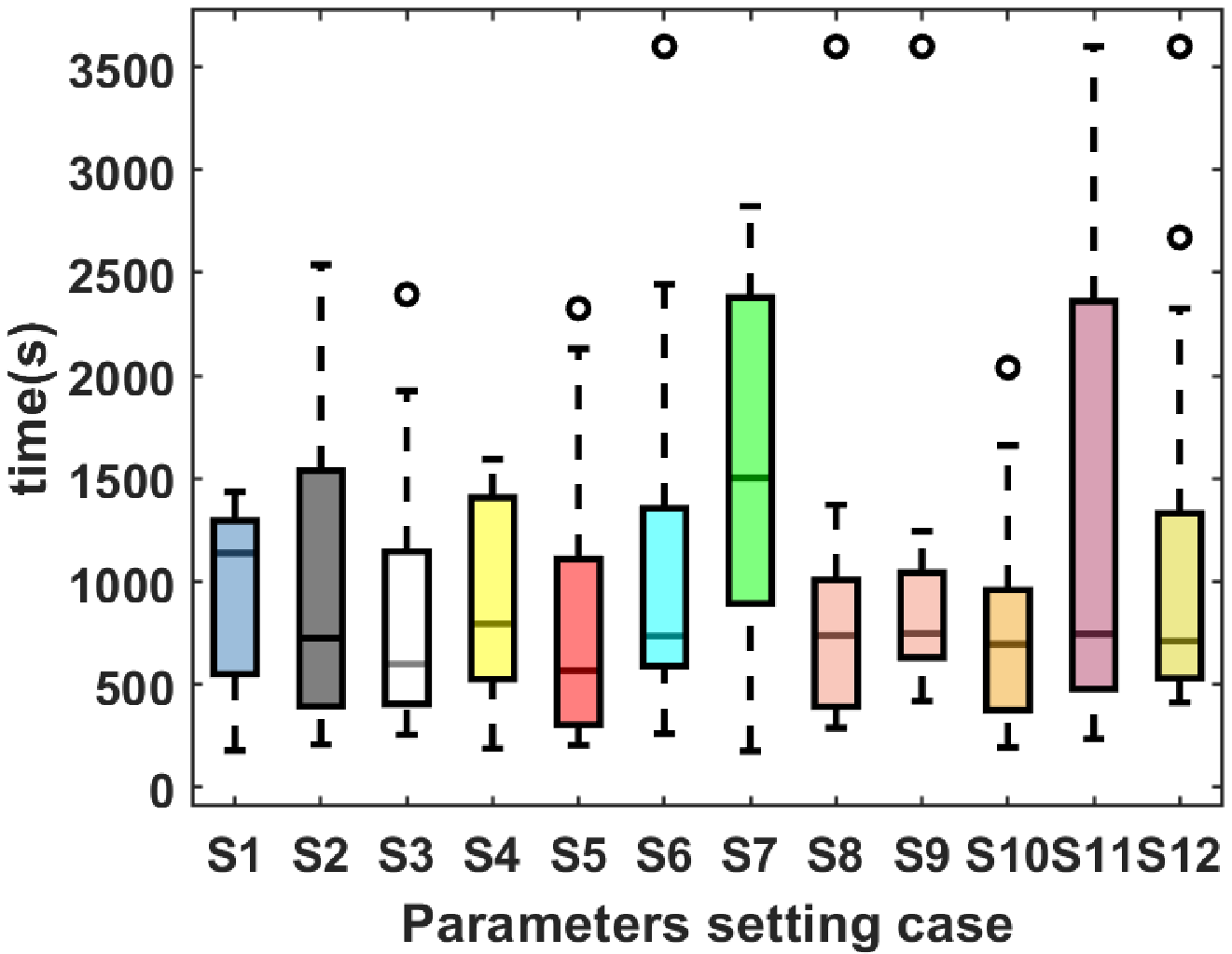}
}\label{parstu-1}
\subfigure[ ]{
\includegraphics[width=5.7cm]{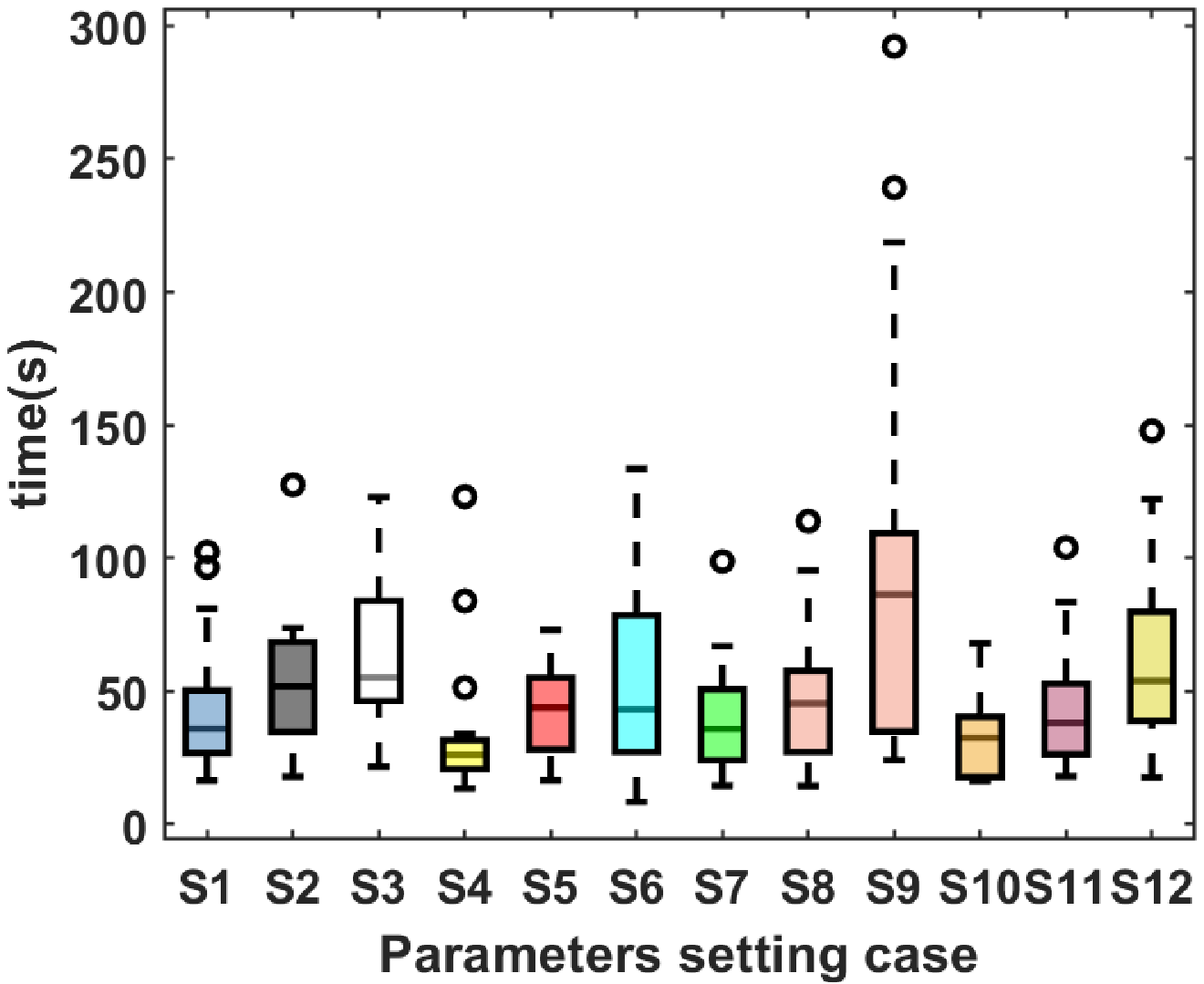}
}\label{parstu-2}
\\
\subfigure[ ]{
\includegraphics[width=5.7cm]{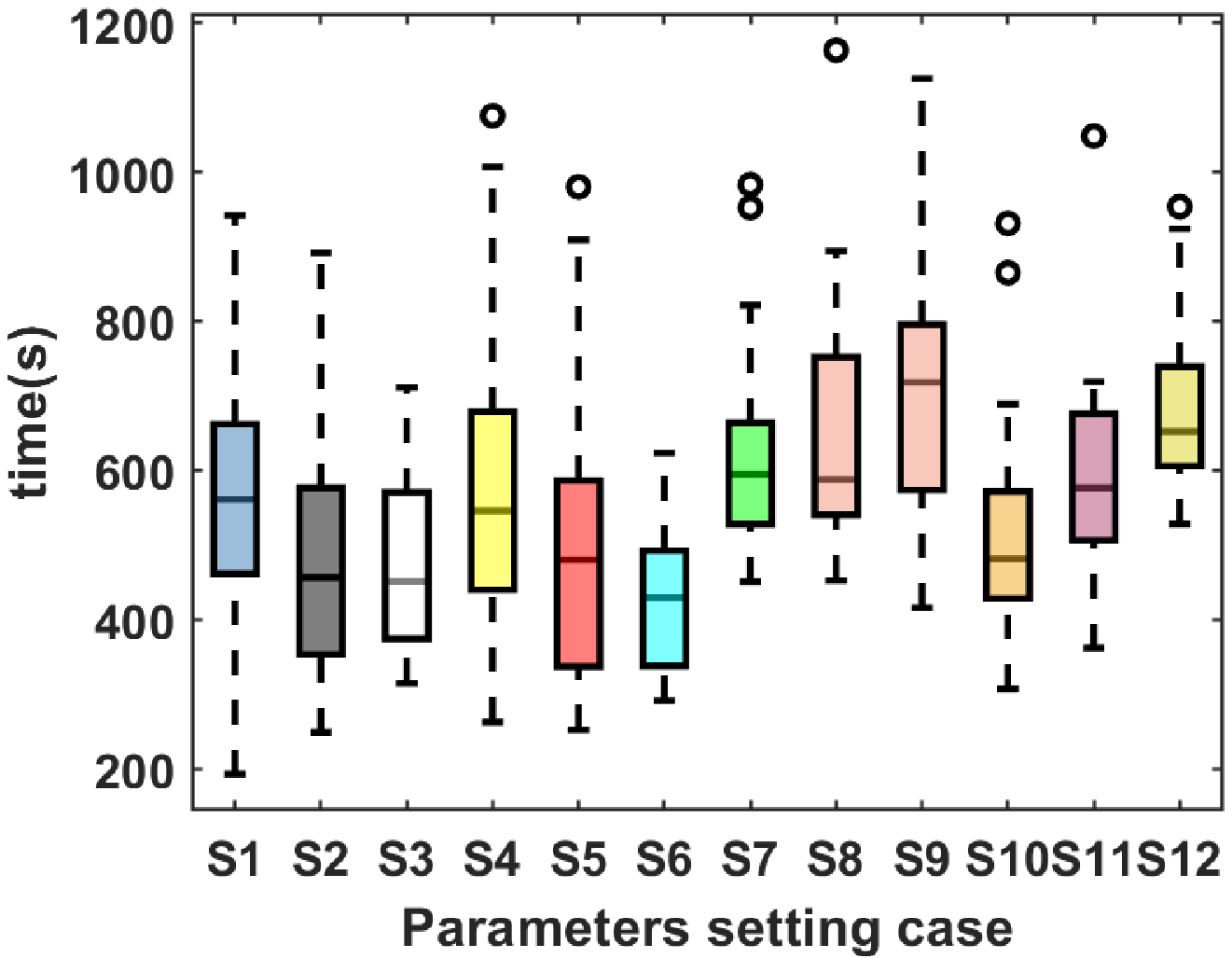}
}\label{parstu-3}
\subfigure[ ]{
\includegraphics[width=5.7cm]{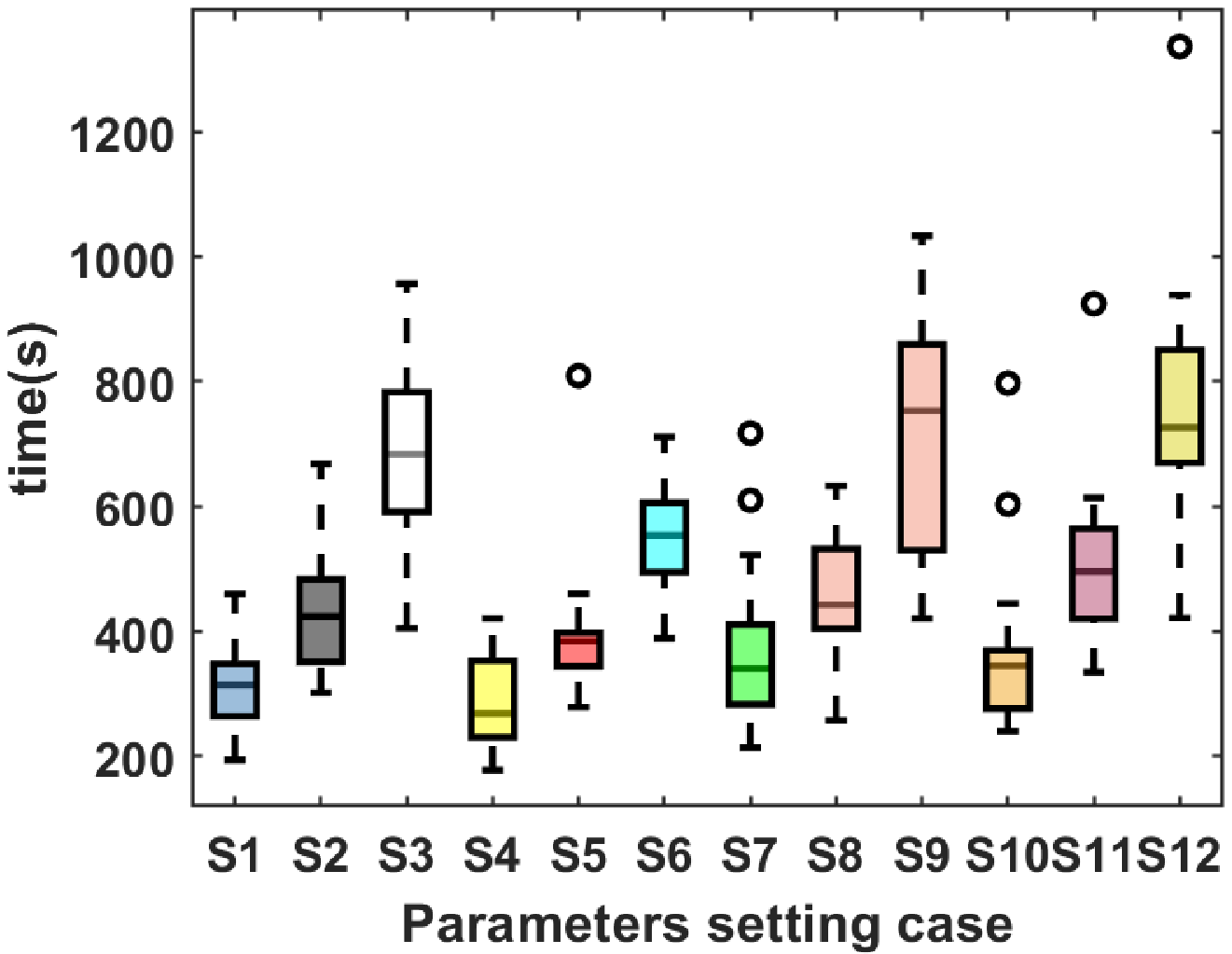}
}\label{parstu-4}
\\
\subfigure[ ]{
\includegraphics[width=5.7cm]{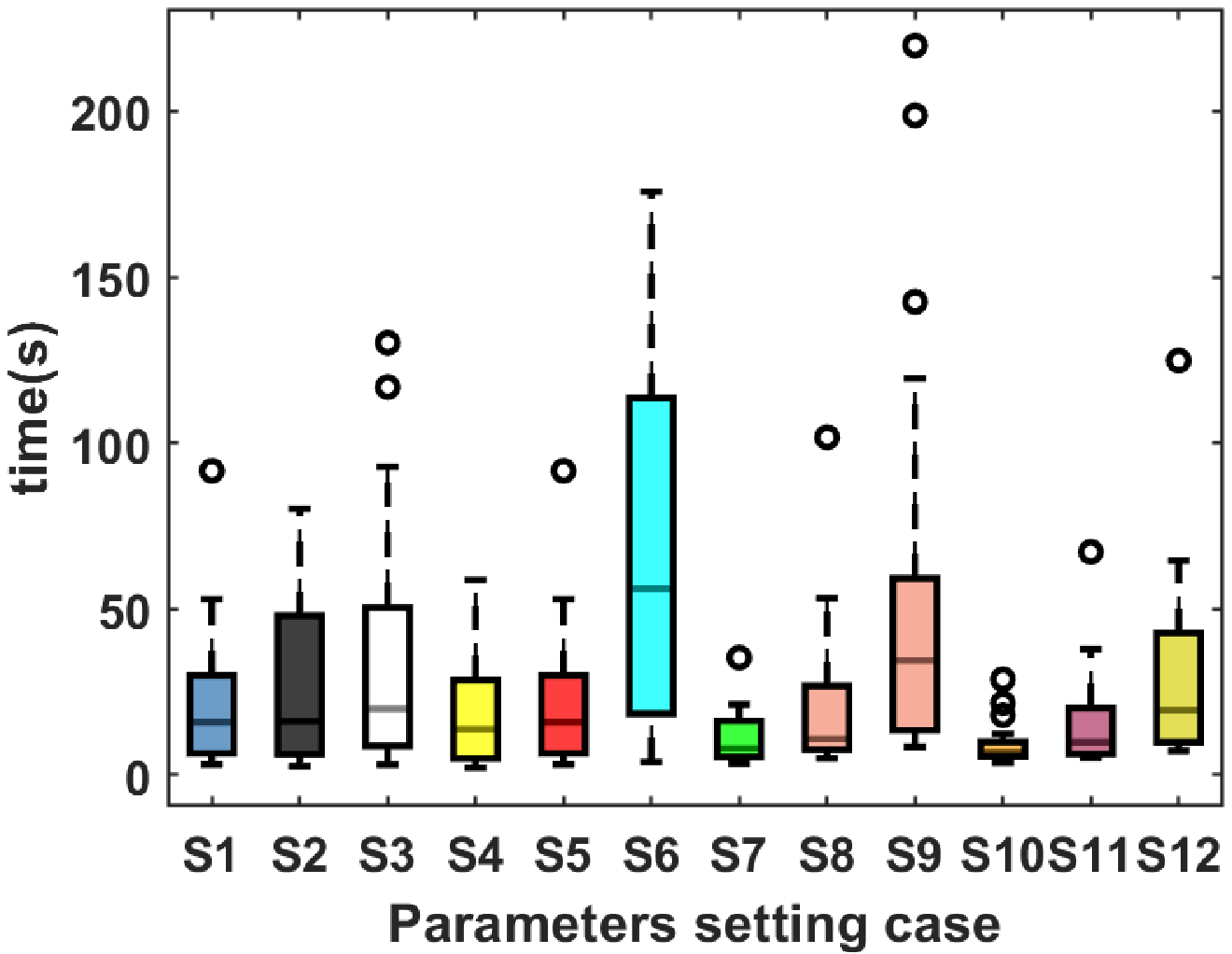}
}\label{parstu-5}
\subfigure[ ]{
\includegraphics[width=5.7cm]{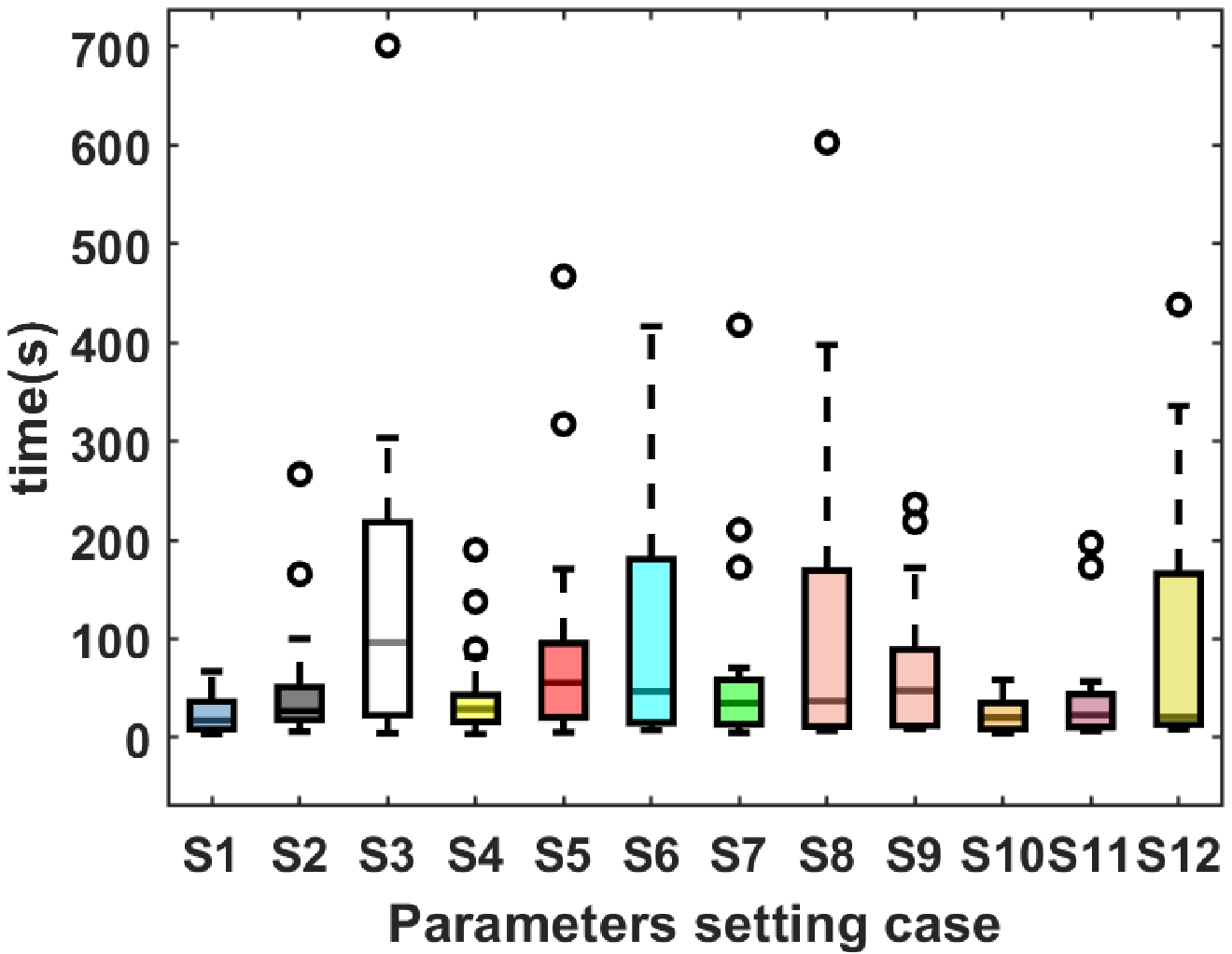}
}\label{parstu-6}
\caption{Statistical results for running time of the DEA-PPM that addresses the $k$-coloring problem of selected benchmark problems. (a)\emph{DSJC500.5}($k=48$). (b)\emph{flat300\_28\_0}($k=31$). (c)\emph{flat1000\_50\_0}($k=50$). (d)\emph{flat1000\_76\_0}($k=86$). (e)\emph{le450\_15c}($k=15$). (f)\emph{le450\_15d}($k=15$).}
\label{parstuFig}
\end{figure}

Besides the one-way ANOVA test, we also present a descriptive comparison for the composite impact of sevaral parameter settings.
With the parameter combinations presented in Tab. \ref{parsetcasTab}, statistical results for running time of $30$ independent runs are included in Fig. \ref{parstuFig}.


\begin{table}[!hbt]
\centering
\caption{Candidate parameter settings of DEA-PPM.}
\resizebox{\textwidth}{!}{
\begin{tabular}{l|cccccccccccc}
\toprule
\toprule
\multirow{2}{*}{Parameter}& \multicolumn{12}{c}{Setting}\\
\cline{2-13}
& $S_1$    & $S_2$     & $S_3$     & $S_4$    & $S_5$     & $S_6$     & $S_7$    & $S_8$     & $S_9$     & $S_{10}$   & $S_{11}$ & $S_{12}$    \\
\hline
$np$          & 4    & 4     & 4     & 4    & 4     & 4     & 8    & 8     & 8     & 8    & 8     & 8     \\
$\alpha$      & 0.1  & 0.1   & 0.1   & 0.2  & 0.2   & 0.2   & 0.1  & 0.1   & 0.1   & 0.2  & 0.2   & 0.2   \\
$Iter_{max}$  & 5000 & 10000 & 20000 & 5000 & 10000 & 20000 & 5000 & 10000 & 20000 & 5000 & 10000 & 20000 \\
\bottomrule
\bottomrule
\end{tabular}\vspace{0cm}
}
\label{parsetcasTab}
\end{table}

It indicates that the parameter setting $S_{10}$ leads to the most promising results of DEA-PPM. Combining it with the setting of other parameters, we get the parameter setting of DEA-PPM presented in Tab. \ref{ParSet}, which is adopted in the following experiments.

\begin{table}[!htb]
  \centering
  \caption{Parameter setting of the DEA-PPM for numerical experiments.}\label{ParSet}
\resizebox{\textwidth}{!}{
  \begin{tabular}{ccl}
    \hline\hline
    Parameter & Setting & Description \\
    \hline
    $np$ & 8 & Population size of $\mathbf Q(t)$ and $\mathbf{P}(t)$; \\
    $\alpha$ & $0.2$ & Regulation parameter in equation (\ref{update11}); \\
    $iter_{max}$ & $5\times10^3$ & Iteration budget for the TS; \\
    $p_0$ & $0.98$  & Parameter for update of $\mathbf Q(t)$;\\
    $\Delta\theta_i$ & $0.05\pi$ & Parameter in equation (\ref{update12}); \\
    $\lambda$ & $0.5$ & Parameter in equation (\ref{update14}); \\
    $r$ &  randomly selected from $\{0.2,0.8\}$ & Parameter in Algorithm \ref{Alg_Gen}; \\
    $P_c$ & $0.4$ & Parameter in Algorithm \ref{Alg_MGPX}; \\
    \hline\hline
  \end{tabular}
  }
\end{table}

\subsection{Experiments on the evolution strategies of probability distribution}
In DEA-PPM, the evolution of distribution population is implemented by the orthogonal exploration strategy and the exploitation strategy. We try to validate the positive effects of these strategies in this section.

\begin{figure}[!hbt]
\centering
\subfigure[Easy benchmark problems]{
\includegraphics[width=5.7cm]{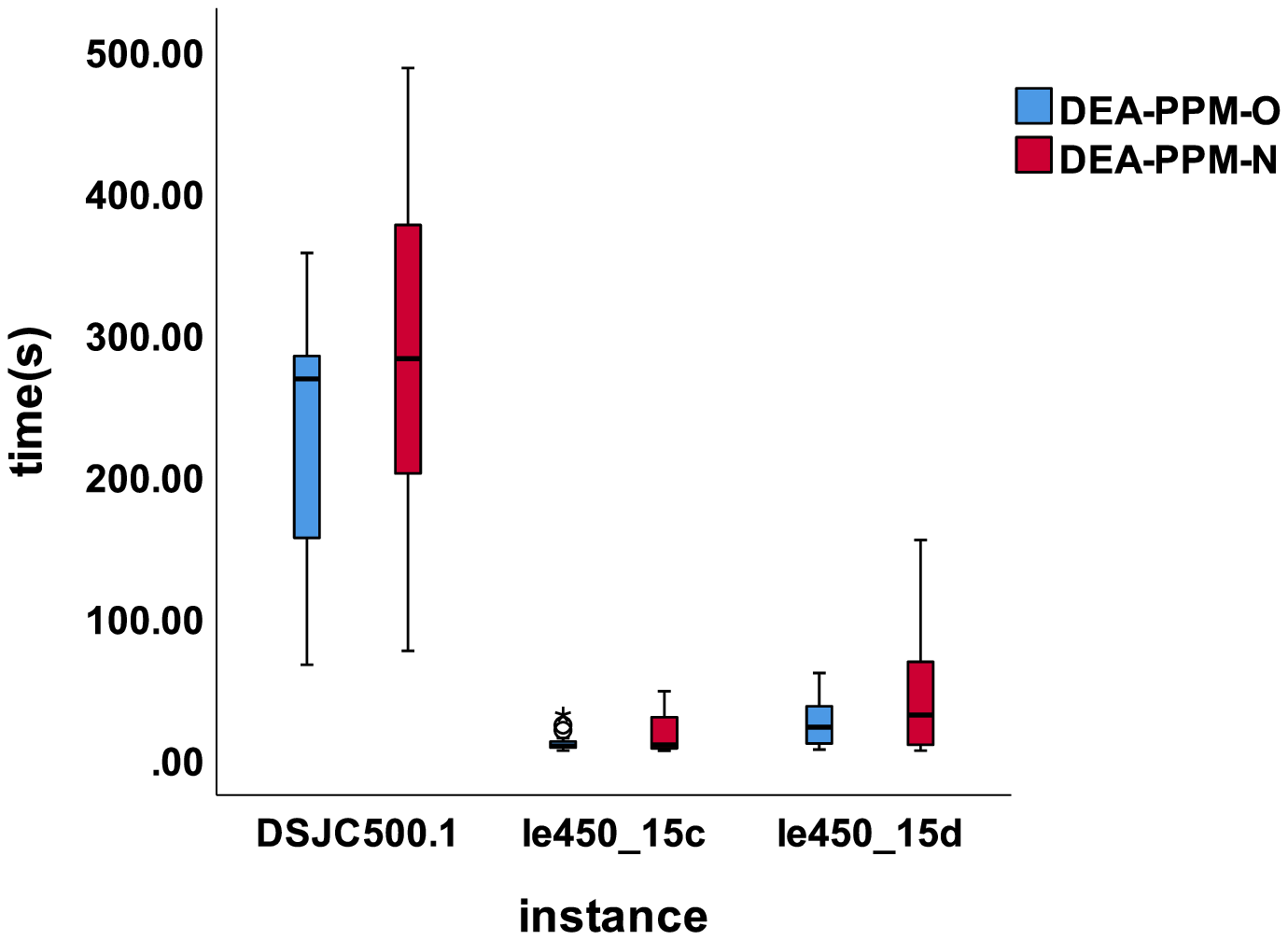}
}\label{or1}
\subfigure[Hard benchmark problems]{
\includegraphics[width=5.7cm]{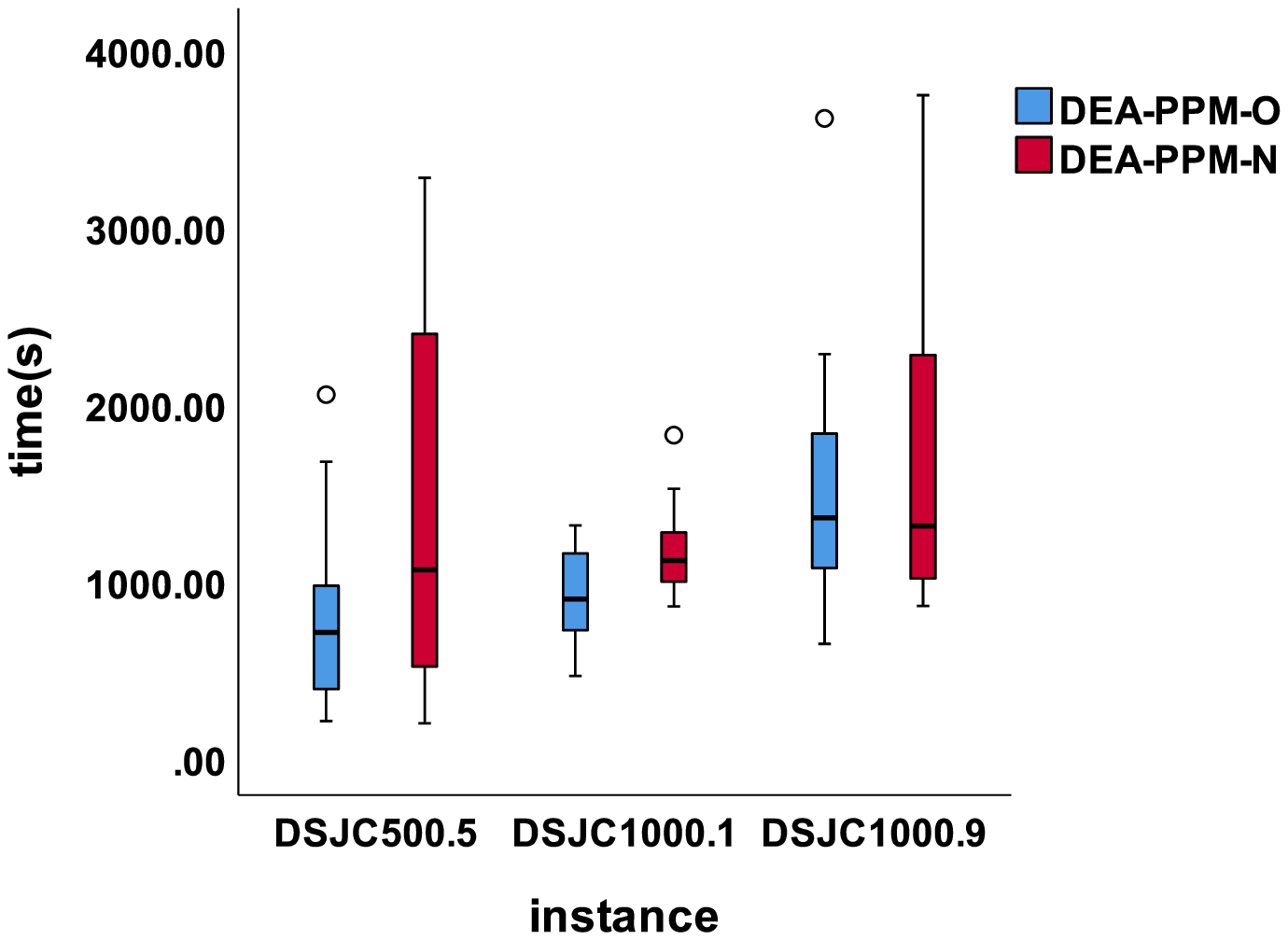}
}\label{or2}
\caption{Comparison of running time between the DEA-PPM-O and the DEA-PPM-N by selected benchmark problems.}
\label{OrthFig}
\end{figure}

To validate the efficiency of the orthogonal exploration strategy, we compare two variants, the DEA-PPM with orthogonal exploration (DEA-PPM-O) and the DEA-PPM without orthogonal exploration (DEA-PPM-N), and show in Fig. \ref{OrthFig} the statistical results of running time for $k$-coloring of the easy benchmark problems (\emph{DSJC500.1} ($k=12$), \emph{le450\_15c} ($k=15$), \emph{led450\_15d} ($k=15$)) and the hard benchmark problems(\emph{DSJC500.5} ($k=48$), \emph{DSJC1000.1} ($k=20$), \emph{DSJC1000.9} ($k=226$)). The box plots imply that with the employment of the orthogonal exploration strategy, DEA-PPM-O performs generally better than DEA-PPM-N, resulting in smaller values of the median value, the quantiles and the standard deviations of running time.

{\color{red}
The positive impact of exploitation strategy is verified by comparing the DEA-PPM with exploitation (DEA-PPM-E) with the variant without exploitation (DEA-PPM-W), and the box plots of running time are included in Fig. \ref{ExplFig}. It is demonstrated that DEA-PPM-E generally outperforms DEA-PPM-W in terms of the median value, the quantiles and the standard deviation of running time.
\begin{figure}[!hbt]
\centering
\subfigure[Easy benchmark problems]{
\includegraphics[width=5.7cm]{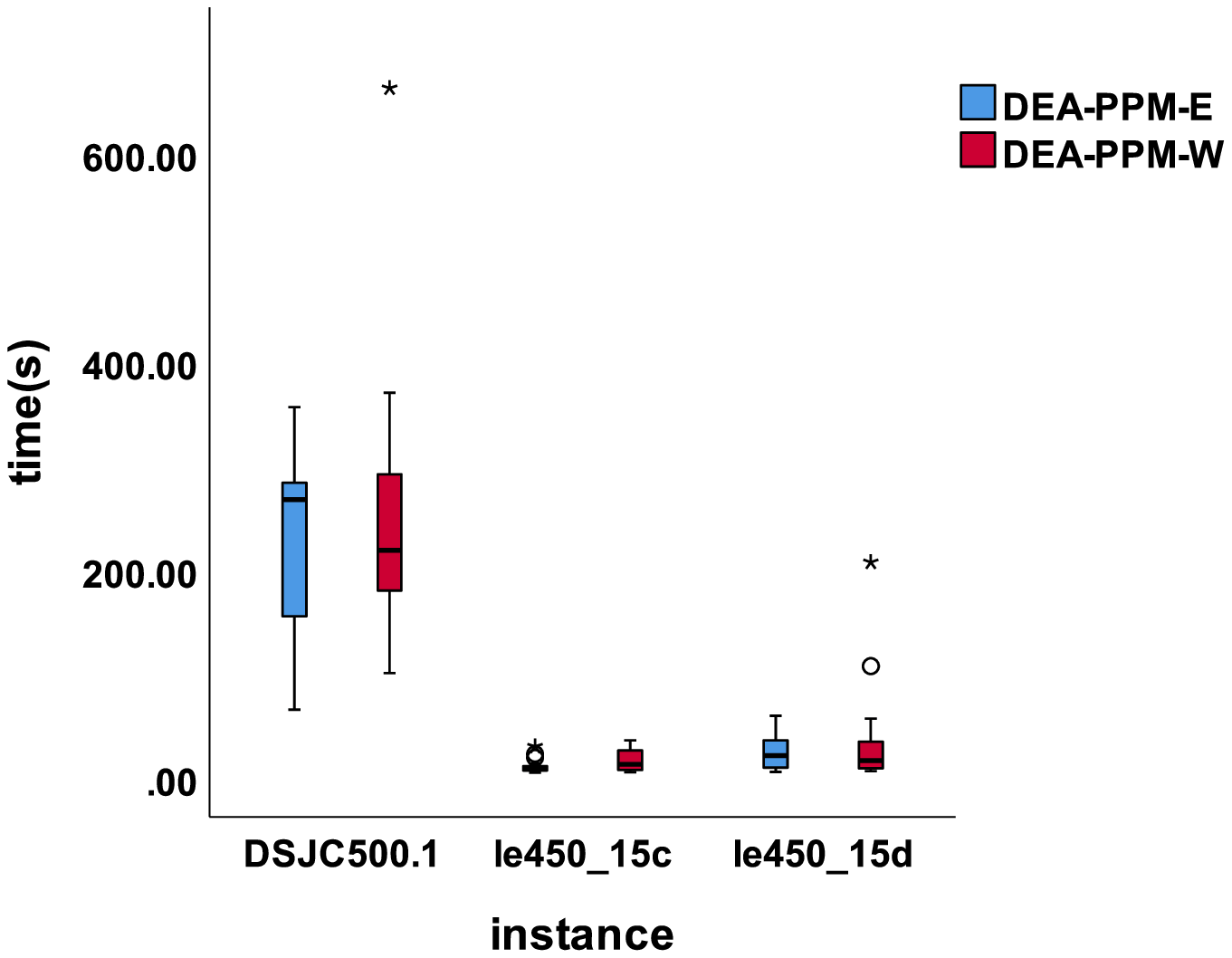}
}\label{or1}
\subfigure[Hard benchmark problems]{
\includegraphics[width=5.8cm]{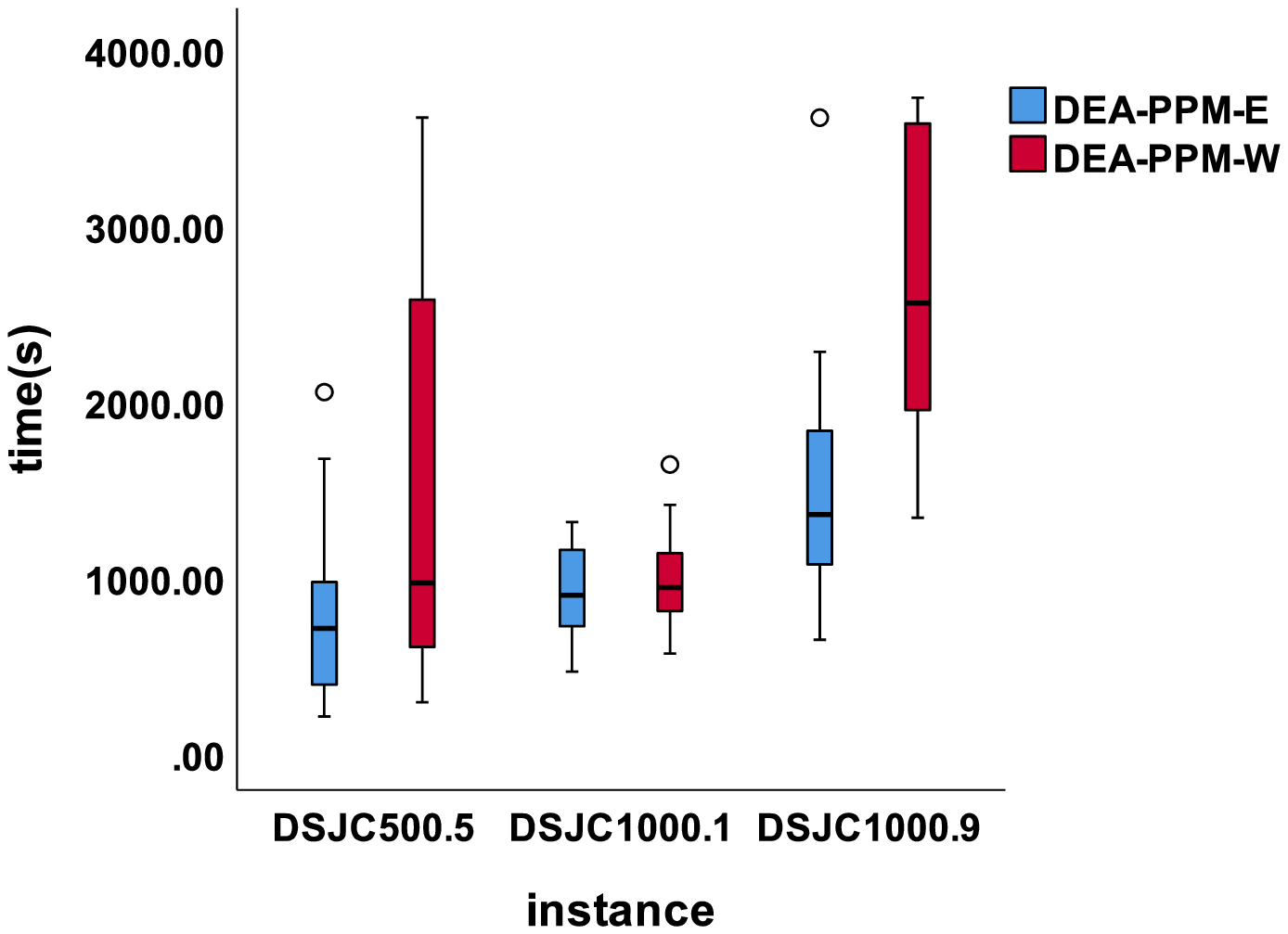}
}\label{or2}
\caption{Comparison of running time between the DEA-PPM-E and the DEA-PPM-W by selected benchmark problems.}
\label{ExplFig}
\end{figure}
}

{\color{blue} Besides the statistical comparison regarding the exact values of running time, we perform a further comparison by the Wilcoxon rank sum test with a significance level of 0.05, where the statistical test is based on the sorted rank of running time instead of its exact values. The results are included in Tab. \ref{tab_s}, where ``P'' is the p-value of hypothesis test. For the test conclusion ``R'', ``+'', ``-'' and ``$\sim$'' indicate that the performance of DEA-PPM is better than, worse than and incomparable to that of the compared variant, respectively. The results demonstrate that DEA-PPM outperforms DEA-PPM-N and DEA-PPM-W on two instances, and is not inferior to them for all benchmark problems. It further validates the conclusion that both the exploration strategy and the exploitation strategy significantly improve the performance of DEA-PPM.

\begin{table}[!htp]
\centering
\caption{Wilcoxon rank-sum test for  the evolution strategies of distribution population.}
{
\begin{tabular}{l|cc|cc}
\hline \hline
\multirow{2}{*}{Instance} & \multicolumn{2}{c|}{DEA-PPM-N} & \multicolumn{2}{c}{DEA-PPM-W}     \\
\cline{2-5}
& P               & R        & P                & R                           \\ \hline
DSJC500.1    & 1.08E-01 & $\sim$ & 9.25E-01     & $\sim$  \\
le450\_15c   & 6.36E-01 & $\sim$ & 4.99E-02 & + \\
le450\_15d   & 2.50E-01 & $\sim$ & 7.76E-01 & $\sim$ \\
DSJC500.5 & 4.97E-02 & + & 5.29E-02 & $\sim$\\
DSJC1000.1 & 6.56E-03 & + & 4.57E-01 & -  \\
DSJC1000.9 & 7.15E-01     & $\sim$ & 4.68E-05 & +  \\

\hline
+/$\sim$/-                                  & \multicolumn{2}{c|}{2/4/0} & \multicolumn{2}{c}{2/4/0} \\
\hline \hline
\end{tabular}
}
\label{tab_s}
\end{table}

}

\subsection{Numerical comparison with the state-of-the-art algorithms}

To demonstrate the competitiveness of DEA-PPM, we perform numerical comparison for the chromatic problem and the $k$-coloring problem with  SDGC~\cite{Galn2017SimpleDG}, MACOL~\cite{lu2010memetic}, SDMA~\cite{Sun2021ASM}, PLSCOL~\cite{Zhou2018ImprovingPL}, and HEAD~\cite{moalic2018variations}, the parameter settings of which are presented in Tab. \ref{para_num}. If an algorithm cannot address the chromatic problem or the $k$-coloring problem in 3600 seconds,  a failed run is recorded by the running time of 3600 seconds.

\begin{table}[!htb]
\centering
  \caption{Parameter settings of the compared algorithms.}\label{para_num}
\resizebox{\textwidth}{!}{
\begin{tabular}{llll}
 \hline\hline
Algorithms                     & Parameters         & Description                                              & Values                \\

                        \hline
SDGC                    & $it$        & The number of iterations;                        & $1\times10^5$  \\

\hline
\multirow{5}{*}{MACOL}  & $p$         & Size of population;                              & $20$               \\
                        & $\alpha$         & Depth of TS;                                     & $1\times10^5$            \\
                        & $m$         & Number of parents for crossover;                 & A random number in $\{2,\dots ,6\}$    \\
                        & $p_{r}$        & Probability for accepting worse offspring;       & $0.2$              \\
                        & $\lambda$         & Parameter for goodness score function;           & $0.08$             \\

                         \hline
\multirow{5}{*}{SDMA}   & $\beta $          & Search depth of weight tabu coloring;            & $1\times10^6$           \\
                        & $tt_{w}$      & Tabu tenure of weight tabu coloring;             & $rand(10) + f^{'}$    \\
                        & $tt$        & Tabu tenure of perturbation;                     & $rand(1000) + f^{'}$ \\
                        & $L$         & Level limit of coarsening phase;                 & $5$                \\
                        & $\lambda$         & Unimproved consecutive rounds for best solution; & $10$               \\

                        \hline
\multirow{6}{*}{PLSCOL} & $\omega $         & Noise probability;                               & $0.2$              \\
                        & $\alpha$         & Reward factor for correct group;                 & $0.1$              \\
                        & $\beta $         & Penalization factor for incorrect group;         & $\left [ 0.05,0.45 \right ]$   \\
                        & $\gamma $         & Compensation factor for expected group;          & $0.3$              \\
                        & $\rho $         & Smoothing coefficient;                           & $0.5$              \\
                        & $p_{0}$        & Smoothing threshold;                             & $0.995$            \\

\hline
\multirow{2}{*}{HEAD}   & $Iter_{TC}$    & Depth of TS;                & $1\times10^5$         \\
                        & $Iter_{cycle}$ & The number of generations into one cycle;        & $10$               \\

    \hline\hline
\end{tabular}
}
\end{table}


\subsubsection{Comparison for the chromatic number problem}

In order to verify the competitiveness of DEA-PPM on the chromatic number problem, we compare it with SDGC, MACOL, SDMA, PLSCOL and HEAD by 8 selected benchmark problems, and the statistical results of 30 independent runs are collected in Tab. \ref{ChrTab}, where  $k_{ave}$, $k_{min}$, $k_{max}$ and $k_{std}$ represent the average color number, the maximum color number, the minimum color number and the standard deviation of color numbers, respectively. The best results are highlighted by bold texts.
\begin{table}[!hbt]
\centering
\caption{Comparison of DEA-PPM with SDGC, MACOL, SDMA, PLSCOL and HEAD for the chromatic number problem.}
\resizebox{\textwidth}{!}{
\begin{tabular}{c|c|l|cccc|c|c|l|cccc}
\hline \hline
Instance    & $\chi(G)$ & Algorithm & $k_{ave}$  & $k_{min}$ & $k_{max}$ & $k_{std}$ & Instance    & $\chi(G)$ & Algorithm & $k_{ave}$  & $k_{min}$ & $k_{max}$ & $k_{std}$  \\
\hline
\multirow{6}{*}{fpsol2\_i\_2}   & \multirow{6}{*}{30}      & SDGC      & 85.4  & 79   & 91   & 3.71 &
\multirow{6}{*}{le450\_15c}     & \multirow{6}{*}{15}    & SDGC     & 29.07 & 28   & 32   & 1.46 \\
                                &                          & MACOL     & 88.3  & 88   & 89   & 0.46  &
                                &                          & MACOL     & 19.4  & 18   & 21   & 1.2  \\
                                &                          & SDMA      & 59.41 & 53   & 73   & 6.22  &
                                &                          & SDMA      & 30.93 & 28   & 38   & 3.12 \\
                                &                          & PLSCOL    & 73    & 71   & 77   & 1.75 &
                                &                          & PLSCOL    & 16.1  & 15   & 17   & 0.4  \\
                                &                          & HEAD      & 74.7  & 71   & 78   & 1.97 &
                                &                          & HEAD      & 15.87 & 15   & 16   & 0.34 \\
                                &                          & DEA-PPM   & \textbf{30}    & \textbf{30}   & \textbf{30}   & \textbf{0} &
                                &                          & DEA-PPM   & \textbf{15}    & \textbf{15}   & \textbf{15}   & \textbf{0}    \\
 \hline
\multirow{6}{*}{fpsol2\_i\_3}   & \multirow{6}{*}{30}      & SDGC      & 85.5  & 79   & 95   & 4.35 &
\multirow{6}{*}{le450\_15d}     & \multirow{6}{*}{15}      & SDGC      & 30.27 & 27   & 34   & 2.24 \\
                                &                          & MACOL     & 88    & 87   & 89   & 0.45 &
                                &                          & MACOL     & 18.7  & 17   & 21   & 1.35 \\
                                &                          & SDMA      & 57.86 & 51   & 65   & 5.28 &
                                &                          & SDMA      & 32.72 & 31   & 38   & 2.17 \\
                                &                          & PLSCOL    & 71.67 & 66   & 77   & 3.47 &
                                &                          & PLSCOL    & 16.33 & 16   & 17   & 0.47 \\
                                &                          & HEAD      & 75.57 & 73   & 78   & 1.54 &
                                &                          & HEAD      & 15.93 & 15   & 16   & 0.25 \\
                                &                          & DEA-PPM   & \textbf{30}    & \textbf{30}   & \textbf{30}   & \textbf{0}    &
                                &                          & DEA-PPM   & \textbf{15}    & \textbf{15}   & \textbf{15}   & \textbf{0}    \\
 \hline
\multirow{6}{*}{flat300\_26\_0} & \multirow{6}{*}{26}      & SDGC      & 40.5  & 38   & 44   & 1.5  &
\multirow{6}{*}{DSJC500\_1}     & \multirow{6}{*}{12}      & SDGC      & 16.67 & 16   & 18   & 0.79 \\

                                &                          & MACOL     & 31.8  & 31   & 32   & 0.4  &
                                &                          & MACOL     & 13    & 13   & 13   & 0    \\

                                &                          & SDMA      & 44.83 & 43   & 51   & 3.12 &
                                &                          & SDMA      & 20.76 & 17   & 29   & 3.39 \\

                                &                          & PLSCOL    & \textbf{26}    & \textbf{26}   & \textbf{26}   & \textbf{0}    &
                                &                          & PLSCOL    & 12.77 & 12   & 13   & 0.42 \\

                                &                          & HEAD      & \textbf{26}    & \textbf{26}   & \textbf{26}   & \textbf{0}    &
                                &                          & HEAD      & 13    & 13   & 13   & 0    \\

                                &                          & DEA-PPM   & \textbf{26}    & \textbf{26}   & \textbf{26}   & \textbf{0}    &
                                &                          & DEA-PPM   & \textbf{12}    & \textbf{12}   & \textbf{12}   & \textbf{0}    \\
 \hline
\multirow{6}{*}{flat300\_28\_0} & \multirow{6}{*}{28}      & SDGC      & 40.83 & 40   & 42   & 0.58 &
\multirow{6}{*}{DSJC1000\_1}    & \multirow{6}{*}{20}      & SDGC      & 31.63 & 31   & 32   & 0.48 \\

                                &                          & MACOL     & 32    & 32   & 32   & 0    &
                                &                          & MACOL     & 80.37 & 76   & 82   & 2.79 \\

                                &                          & SDMA      & 45.72 & 43   & 54   & 3.64 &
                                &                          & SDMA      & 86.21 & 73   & 94   & 5.55 \\

                                &                          & PLSCOL    & \textbf{31}    & \textbf{30}   & 32   & 0.73 &
                                &                          & PLSCOL    & \textbf{21}    & \textbf{21}   & \textbf{21}   & \textbf{0}    \\

                                &                          & HEAD      & \textbf{31}    & {31}   & \textbf{31}   & \textbf{0}    &
                                &                          & HEAD      & \textbf{21}    & \textbf{21}   & \textbf{21}   & \textbf{0}   \\

                                &                          & DEA-PPM   & \textbf{31}    & {31}   & \textbf{31}   & \textbf{0}    &
                                &                          & DEA-PPM   & \textbf{21}    & \textbf{21}   & \textbf{21}   & \textbf{0}    \\

\hline \hline
\end{tabular}
}
\label{ChrTab}
\end{table}

{\color{red}It is shown that DEA-PPM generally outperforms the other five state-of-the-art algorithms on $k_{ave}$, $k_{min}$, $k_{max}$ and $k_{std}$ of 30 independent runs. Attributed to the population-based distribution evolution strategy, the global exploration ability of DEA-PPM is enhanced significantly. Moreover, the inherited initialization strategy improve the searching efficiency of the inner loop for search of $k$-coloring assignment. As a result, it can address these problems efficiently and obtain $\chi(G)$ with a 100\% success rate for all of eight selected problems.

It is noteworthy that the competitiveness is partially attributed to the IVR strategy introduced by DEA-PPM, especially for the sparse benchmark graphs \emph{fpsol2.i.2} and \emph{fpsol2.i.3}. Numerical implementation shows that when $k=30$,  introduction of the IVR strategy reduces the vertex number of \emph{fpsol2.i.2} and \emph{fpsol2.i.3} from 451 and 425 to 90 and 88, respectively. Thus, the scale of the reduced graph $G'$ is significantly cut down for \emph{fpsol2.i.2} and \emph{fpsol2.i.3}, which greatly improves the efficiency of the $k$-coloring process validated by the inner loop of DEA-PPM.}

{\color{red} However, it demonstrates that DEA-PPM, PLSCOL and HEAD get consistent results on the instances \emph{flag300\_26\_0} and \emph{DSJC1000\_1}, and the best results of DEA-PPM and HEAD is a bit worse than that of PLSOCL. Accordingly, we further compare their performance by the Wilcoxon rand sum test. If the compared algorithms obtain different results of color number, the sorted rank is calculated according to the color number; while they get consistent results of color number, the rank sum test is performed according to the running time of 30 independent runs.
\begin{table}[!htp]
\centering
\caption{Wilcoxon rank sum test for the comparison of performance on the chromatic number problem.}
\resizebox{\textwidth}{!}{
\begin{tabular}{l|cc|cc|cc|cc|cc}
\hline \hline
\multirow{2}{*}{Instance}        & \multicolumn{2}{c|}{SDGC}  & \multicolumn{2}{c|}{MACOL} & \multicolumn{2}{c|}{SDMA} &\multicolumn{2}{c|}{PLSCOL} & \multicolumn{2}{c}{HEAD} \\
\cline{2-11}
    & P & R & P & R & P & R & P & R & P & R\\
    \hline
    fpsol2\_i\_2 &	1.09E-12 &	+ &	2.90E-13 &	+ &	1.13E-12 &	+ &	9.96E-13 &	+ &	1.11E-12 &	+\\
fpsol2\_i\_3 &	1.17E-12 &	+ &	1.59E-13 &	+ &	1.13E-12 &	+ &	1.12E-12 &	+ &	1.03E-12 &	+ \\
flat300\_26\_0 &	7.98E-13 &	+ &	1.55E-13 &	+ &	1.11E-12 &	+ &	 4.81E-11 &	- &	6.73E-01 &	$\sim$ \\
flat300\_28\_0 &	4.27E-13 &	+ &	1.69E-14 &	+ &	1.05E-12 &	+ &	6.31E-01 &	$\sim$ &	0.028 &	+ \\
le450\_15c &	8.93E-13 &	+ &	7.31E-13 &	+ &	1.13E-12 &	+ &	2.05E-10 &	+ &	1.97E-11 &	+ \\
le450\_15d &	1.05E-12 &	+ &	5.16E-13 &	+ &	9.63E-13 &	+ &	3.37E-13 &	+ &	7.15E-13 &	+ \\
DSJC500\_1 &	6.21E-13 &	+ &	1.69E-14 &	+ &	9.31E-13 &	+ &	1.47E-09 &	+ &	1.69E-14 &	+ \\
DSJC1000\_1 &	3.80E-13 &	+ &	1.09E-12 &	+ &	1.15E-12 &	+ &	2.74E-11 &	- &	3.73E-09 &	- \\

\hline
+/$\sim$/-    & \multicolumn{2}{c|}{8/0/0} & \multicolumn{2}{c|}{8/0/0} & \multicolumn{2}{c|}{8/0/0} & \multicolumn{2}{c|}{5/1/2} & \multicolumn{2}{c}{6/1/1}\\
\hline \hline
\end{tabular}
}
\label{cTab}
\end{table}

The test results demonstrate that DEA-PPM does outperform SDGC, MACOL and SDMA on the selected benchmark problems. For the instance \emph{flat300\_26\_0}, it is shown in Tab. \ref{ChrTab} that DEA-PPM, PLSCOL and HEAD can address the chromatic number in 3600s. While the running time is compared by the rank sum test, we get the conclusion that PLSCOL runs fast than DEA-PPM. Considering the instance \emph{DSJC1000\_1}, DEA-PPM, PLSCOL and HEAD stagnate at the assignment of 21 colors. However, the rank sum test shows that DEA-PPM is inferior to PLSCOL and HEAD in term of the running time.
}

\subsubsection{Comparison  for the $k$-coloring problem}

Numerical results on the chromatic problem imply that DEA-PPM, PLSCOL and HEAD outperform SDGC, MACOL and SDMA, but the superiority of DEA-PPM over PLSCOL and HEAD is dependent on the benchmark instances. To further compare DEA-PPM with PLSCOL and HEAD, we investigate their performance for the $k$-coloring problem, where $k$ is set as the chromatic number of the investigate instance.
For 18 selected benchmark problems collected in Tab. \ref{kTab}, we present the success rate (SR) and average runtime (T) of 30 independent runs, and the best results are highlighted by bold texts.

\begin{table}[!htp]
\centering
\caption{Numerical results of DEA-PPM, HEAD and PLSCOL for the $k$-coloring problem}
\resizebox{\textwidth}{!}{
\begin{tabular}{l|c|cc|cc|cc}
\hline \hline
\multirow{2}{*}{Instance}        & \multirow{2}{*}{$k$}  & \multicolumn{2}{c|}{PLSCOL} & \multicolumn{2}{c|}{HEAD} & \multicolumn{2}{c}{DEA-PPM} \\
\cline{3-8}
                &        & SR     & T(s)        & SR    & T(s)       & SR      & T(s)     \\
\hline
DSJC125.5       & 17        & \textbf{30/30}  & \textbf{0.31}           & \textbf{30/30} & 0.55          & \textbf{30/30}   & 0.65     \\
DSJC125.9       & 44        & \textbf{30/30}  & \textbf{0.08}          & \textbf{30/30} & 0.12          & \textbf{30/30}   & 0.39     \\
DSJC250.5       & 28        & \textbf{30/30}  & \textbf{17.60}        & \textbf{30/30} & 37.03        & \textbf{30/30}   & 23.93    \\
DSJC250.9       & 72        & \textbf{30/30}  & \textbf{6.22}           & \textbf{30/30} & 7.20         & \textbf{30/30}   & 42.97    \\
DSJC500.1       & 12        & 9/30   & 3140.21       & 29/30 & 1655.11       & \textbf{30/30}   & \textbf{226.14}   \\
DSJC500.5       & 48      & 0/30   & 3600           & \textbf{30/30} & 1176.30       & \textbf{30/30}   & \textbf{771.39}   \\
DSJC500.9       & 126      & 0/30   & 3600          & 1/30  & 3504.49      & \textbf{3/30}    & \textbf{3346.64}  \\
DSJC1000.1      & 20        & 0/30   & 3600           & 0/30  & 3600         & \textbf{30/30}   & \textbf{902.53}   \\
DSJC1000.5      & 85       & 0/30   & 3600          & \textbf{30/30} & 2575.73      & 23/30   & \textbf{2271.70}  \\
DSJC1000.9      & 225      & 0/30   & 3600         & \textbf{19/30} & \textbf{2784.67}      & 12/30   & 3240.40  \\
le450\_15c      & 15        & 0/30   & 3600           & \textbf{30/30} & 400.85        & \textbf{30/30}   & \textbf{9.34}     \\
le450\_15d      & 15       & 0/30   & 3600          & 27/30 & 1121.86       & \textbf{30/30}   & \textbf{24.60}    \\
flat300\_20\_0  & 20        & \textbf{30/30}  & \textbf{0.11}           & \textbf{30/30} & 0.20          & \textbf{30/30}   & 0.81     \\
flat300\_26\_0  & 26        & \textbf{30/30}  & \textbf{3.47}          & \textbf{30/30} & 8.81         & \textbf{30/30}   & 15.46    \\
flat300\_28\_0  & 30       & \textbf{5/30}   & \textbf{3196.66}        & 0/30  & 3600    & 0/30    & 3600     \\
flat1000\_50\_0 & 50       & \textbf{30/30}  & \textbf{159.32}         & \textbf{30/30} & 433.27        & \textbf{30/30}   & 636.96   \\
flat1000\_60\_0 & 60        & \textbf{30/30}  & \textbf{347.74}         & \textbf{30/30} & 580.71        & \textbf{30/30}   & 843.81   \\
flat1000\_76\_0 & 84        & 0/30   & 3600        & 23/30 & 2834.23   & \textbf{30/30}   & \textbf{2139.33}  \\
\hline
\multicolumn{2}{c|}{Average Rank} & 1.94 & 2 & 1.38 & 2.05 & \textbf{1.16} & \textbf{1.83}\\
\hline \hline
\end{tabular}
}
\label{kTab}
\end{table}

{\color{blue} Thanks to the incorporation of the population-based distribution evolution strategy, the global exploration of DEA-PPM has been significantly improved, resulting in better success rate for most of the selected problems except for  \emph{DSJC1000.9} and \emph{flag\_300\_28\_0}. Accordingly, the average rank of DEA-PPM is 1.16, better than 1.94 of PLSCOL and 1.38 of HEAD. The global exploration improved by the population-based distribution strategy and the IVR contributes to faster convergence of DEA-PPM for the complicated benchmark problems, however,  increases the generational complexity of DEA-PPM, which leads to its slightly increased running time in some small-scale problems.  Consequently, DEA-PPM gets the first place with the average running-time rank 1.83.}

%

{\color{red}Further investigation of the performance is conducted by the Wilcoxon rank sum test of running time. With a significance level of 0.05, the results are presented in Tab. \ref{wilTab}, where ``P'' is the p-value of hypothesis test.  While both HEAD and DEA-PPM cannot get legal color assignments for \emph{flat300\_28\_0}, the Wilcoxon rank sum test is conducted by the numbers of conflicts of 30 independent runs.

It is shown that DEA-PPM performs better than PLSCOL for 2 of 9 selected instances with vertex number less than 500, and better than HEAD for 3 of 9 problems, but performs a bit worse than PLSCOL and HEAD for most of small-scale instances. However, It outperforms PLSCOL and HEAD on the vast majority of instances with vertex number greater than or equal to 500. Therefore, we can conclude that DEA-PPM is competitive to PLSCOL and HEAD on large-scale GCPs, which is  attributed to the composite function of the population-based distribution evolution mechanism and the IVR strategy.

\begin{table}[!htp]
\centering
\caption{Results of Wilcoxon rank-sum test for performance comparison.}
\resizebox{\textwidth}{!}{
\begin{tabular}{l|cc|cc|l|cc|cc}
\hline \hline
\multirow{2}{*}{Instance ($n<500$)} & \multicolumn{2}{c}{PLSCOL} & \multicolumn{2}{|c|}{HEAD}    & \multirow{2}{*}{Instance ($n\ge 500$)} & \multicolumn{2}{c|}{PLSCOL}                  & \multicolumn{2}{c}{HEAD}                    \\
\cline{2-5} \cline{7-10}
                                           & P               & R        & P                & R        &                                  & P                    & R                    & P                    & R                    \\ \hline
DSJC125.5    & 4.00E-03 & - & 0.46     & $\sim$  & DSJC500.1     & 1.96E-10 & + & 1.10E-11 & + \\
DSJC125.9    & 3.01E-11 & - & 3.00E-03 & - & DSJC500.5     & 1.21E-12 & + & 2.00E-03   & + \\
DSJC250.5    & 0.22     & $\sim$ & 0.38     & $\sim$ & DSJC500.9     & 0.08     & $\sim$ & 0.34   & $\sim$ \\
DSJC250.9    & 6.01E-08 & - & 6.53E-08 & - & DSJC1000.1    & 1.21E-12 & + & 1.21E-12 & + \\
le450\_15c   & 5.05E-13 & + & 1.40E-11 & + & DSJC1000.5    & 5.85E-09 & + & 0.06   & $\sim$ \\
le450\_15d   & 5.05E-13 & + & 1.40E-11 & + & DSJC1000.9    & 1.53E-04 & + & 0.03   & + \\
flat300\_20\_0 & 3.01E-11 & - & 3.01E-11 & -& flat1000\_50\_0 & 3.02E-11 & - & 1.75E-05 & - \\
flat300\_26\_0 & 8.15E-11 & - & 1.39E-06 & - & flat1000\_60\_0 & 8.99E-11 & - & 6.36E-05 & - \\
flat300\_28\_0 & 0.02     & - & 4.62E-05 & +  & flat1000\_76\_0 & 3.45E-07 & + & 0.02   & + \\
\hline
+/$\sim$/-                                  & \multicolumn{2}{c|}{2/1/6} & \multicolumn{2}{c|}{3/2/4} & +/$\sim$/-                        & \multicolumn{2}{c|}{6/1/2}                  & \multicolumn{2}{c}{5/2/2}\\
\hline \hline
\end{tabular}
}
\label{wilTab}
\end{table}
}

\section{Conclusion and Future Work}\label{con}
  {\color{red} To address the graph coloring problems efficiently, this paper develops a distribution evolution algorithm based on a population of probability model (DEA-PPM). Incorporating the merits of the respective probability models in EDAs and QEAs, we introduce a novel distribution model, for which an orthogonal exploration strategy is proposed to explore the probability space efficiently. Meanwhile, an inherited initialization is employed to accelerate the process of color assignment.}
  \begin{itemize}
    \item Assisted by an iterative vertex removing strategy and a TS-based local search process, DEA-PPM can achieve excellent performance with small populations, which contributes to its competitiveness on the chromatic problem.
    \item Since the population-based evolution leads to slightly increased generational time complexity of DEA-PPM, its running time for the small-scale $k$-coloring problems is a bit higher than that of the individual-based PLSCOL and HEAD.
    \item DEA-PPM achieves overall outperformance on benchmark problems with vertex numbers greater than 500, because its enhanced global exploration improves the ability of escaping from the local optimal solutions.
    \item The iterative vertex removal strategy reduces sizes of the graphs to be colored, which likewise improves the coloring performance of DEA-PPM.
  \end{itemize}

The proposed DEA-PPM could be extended to other complex problems.
To further improve the efficiency of DEA-PPM, our future work will focus on the adaptive regulation of population size, and the local exploitation is anticipated to be enhanced by utilizing the mathematical characteristics of graph instance. Moreover, we will try to develop a general framework of DEA-PPM to address a variety of combinatorial optimization problems.

\section*{Acknowledgement}
This research was supported in part by the National Key R\& D Program of China [grant number 2021ZD0114600],
in part by the Fundamental Research Funds for the Central
Universities [grant number WUT:2020IB006], and in part by the National Nature Science Foundation of China [grant number 61763010] as well as the Natural Science Foundation of Guangxi [grant number 2021GXNSFAA075011].
\bibliography{mybibfile}

\end{document}